\documentclass[preprint,12pt]{elsarticle}

\usepackage{amssymb}
\usepackage{amsthm}

\usepackage{url} 
\usepackage{booktabs}
\usepackage{comment}
\usepackage{amsmath}
\usepackage{xcolor}
\usepackage{algpseudocode}
\usepackage{algorithm}
\usepackage{multirow}
\theoremstyle{definition}

\usepackage{placeins}

\definecolor{lightgrey}{rgb}{0.85, 0.85, 0.85}

\usepackage{hyperref}

\usepackage{tabularx}
\usepackage{ragged2e} 
\usepackage{booktabs} 
\newcolumntype{C}{>{\Centering\arraybackslash}X}

\usepackage{pifont}
\newcommand{\cmark}{\ding{51}}  
\newcommand{\xmark}{\ding{55}}  

\newcommand{\ent}[1]{\begin{footnotesize}\textsf{#1}\end{footnotesize}}
\newcommand{\fun}[1]{\begin{small}\textsc{#1}\end{small}}

\journal{Applied Soft Computing}

\begin{document}

\begin{frontmatter}



\title{Truthful Text Sanitization Guided by Inference Attacks}


\author[inst1]{Ildikó Pilán}

\author[inst2]{Benet Manzanares-Salor\footnote{Corresponding author. Email: benet.manzanares@urv.cat}}

\author[inst2]{David Sánchez}

\author[inst1]{Pierre Lison}

\affiliation[inst1]{organization={Norwegian Computing Center},
            addressline={Postboks 114 Blindern}, 
            city={Oslo},
            postcode={NO-0314}, 
            country={Norway}}

\affiliation[inst2]{organization={Department of Computer Engineering and Mathematics. CYBERCAT},
            addressline={Universitat Rovira i Virgili}, 
            city={Tarragona},
            postcode={43007}, 
            country={Spain}}

\begin{abstract}
Text sanitization aims to rewrite parts of a document to prevent disclosure of personal information. The central challenge of text sanitization is to strike a balance between privacy protection (avoiding the leakage of personal information) and utility preservation (retaining as much as possible of the document's original content). To this end, we introduce a novel text sanitization method based on \emph{generalizations}, that is, broader but still informative terms that subsume the semantic content of the original text spans. The approach relies on the use of instruction-tuned large language models (LLMs) and is divided into two stages. Given a document including text spans expressing personally identifiable information (PII), the LLM is first applied to obtain truth-preserving replacement candidates for each text span and rank those according to their abstraction level. Those candidates are then evaluated for their ability to protect privacy by conducting \textit{inference attacks} with the LLM. Finally, the system selects the most informative replacement candidate shown to be resistant to those attacks. This two-stage process produces replacements that effectively balance privacy and utility.

We also present novel metrics to evaluate these two aspects without needing to manually annotate documents. Results on the Text Anonymization Benchmark show that the proposed approach, implemented with Mistral 7B Instruct, leads to enhanced utility, with only a marginal ($ < 1$~p.p.) increase in re-identification risk compared to fully suppressing the original spans. Furthermore, our approach is shown to be more truth-preserving than existing methods such as Microsoft Presidio's synthetic replacements. 
\end{abstract}


\end{frontmatter}


\section{Introduction}
\label{sec:intro}

Text documents play a key role in many sectors, including healthcare, legal services and public administrations. However, those documents often contain personal data that, if released without adequate protection, can pose significant privacy risks to the individuals they concern. Regulations such as the General Data Protection Regulation (GDPR) \cite{GDPR} impose strict constraints on how personal information should be protected. In particular, personal information cannot be shared with third parties or employed for secondary purposes without a proper legal ground, such as the explicit consent of the individuals to whom the data refers.

One strategy to overcome those constraints is to \textit{anonymize} the data, in which case the data is no longer considered personal, and thus outside the scope of those regulations \cite{GDPR}. Beyond regulatory compliance, anonymization also helps balance the pursuit of data-driven research and innovation with the protection of individual rights. By reducing the risk of re-identification, anonymization fosters greater trust between individuals and institutions, encouraging more open data sharing practices that ultimately benefit society as a whole.

Multiple text anonymization techniques have been proposed in the literature \cite{lison-text-anonymization-sota}. In particular, \emph{text sanitization} prevents the disclosure of personal information by: (i) \emph{detecting} spans that contain personally identifiable information (PII), and (ii) \emph{masking} those spans. Most works have focused on the first step (PII detection), and very often rely on simple masking techniques, such as removing PIIs or replacing them by generic semantic categories such as \ent{PERSON} or \ent{LOCATION}) \cite{johnson-2020-ner-bert-medical, yang-2015-ner-detect-phi, yogarajan-2018-ner-survey-clinical, huang-2020-ner-texthide, mamede-2016-ner, bubeck-2023-ner-gpt-sparks, ppdp-anthi-bootstrapping}. However, this relatively crude level of protection can undermine the downstream utility of the sanitized texts, which is the ultimate goal of data sharing. As the data privacy literature has highlighted \cite{hundepool-sdc-book}, data protection should aim at balancing the trade-off between the attained level of privacy and the preservation of data utility. For textual data, this means designing effective \emph{text sanitization} mechanisms that preserve the original text's meaning as much as possible while satisfying the privacy needs of the data release.

In this context, large language models (LLMs), which have shown strong capabilities in a wide range of natural language understanding and generation tasks, have emerged as promising tools for enhancing text sanitization. Recent studies have demonstrated their effectiveness in text editing \cite{shu2024rewritelm} and simplification \cite{kew-etal-2023-bless} with strong zero- and few-shot performance due to instruction tuning \citep{Minaee2024-fi}.

Despite these advances, current text sanitization approaches rarely leverage LLMs to \textit{generate} replacements for detected PII spans. Most existing methods rely on suppression or named entity-based substitutions, which fail to produce coherent and realistic text. As a result, there is currently no principled method for generating \emph{truthful and utility-preserving replacements} of sensitive text spans. Addressing this gap is essential for enabling safe and effective data sharing under privacy regulations while maintaining the semantic integrity and analytical value of textual data.

\subsection{Contributions and plan}

In this paper, we propose a novel text sanitization approach, named INference-guided Truthful sAnitization for Clear Text (INTACT). INTACT leverages instruction-tuned open-weights LLMs to replace personal information with less specific, but truth-preserving \emph{generalizations} that retain the original semantic content to the greatest extent possible, while preventing inference attacks from recovering the initial data. 

Our work brings the following contributions:
\begin{itemize}
    \item Unlike previous privacy-oriented text rewriting methods \cite{arranz-etal-2022-mapa,dou-etal-2024-reducing}, the presented method adopts a two-stage strategy. Given a text span to replace, we first (1) generate a sorted list of generalization candidates at different abstraction levels, and then (2) select the most suitable replacement among those alternatives. This selection of the best replacement is done by testing whether the LLM can guess the original span from the replacement and the rest of the text (\emph{inference attacks}). LLMs have indeed  shown to be effective adversaries in anonymization tasks in previous studies, either on their own \cite{patsakis2023man} or augmented with a retrieval component \citep{Charpentier2025-wa}.
    The combination of those two steps not only allows for balancing and customizing the privacy-utility trade-off, but also enables a more granular evaluation of the rewriting process. 
    \item  We present an evaluation framework for text sanitization that measures privacy and utility without reference annotations. This framework includes a novel utility metric, Text Preserved Similarity (TPS), which measures the similarity between original and sanitized documents according to both information content and semantic similarity.
    \item We evaluate and compare our method against several baselines on the Text Anonymization Benchmark (TAB) \cite{pilan-tab}, a dataset explicitly tailored for the evaluation of text sanitization methods. Empirical results show that our method achieves the best privacy-utility trade-off, while keeping the protected data truthful to their original content.
\end{itemize}

As text sanitization is by definition a process that involves the processing of (sometimes highly sensitive) personal data, our approach relies on the use of open-weights LLMs, as those can be run locally without necessitating the transfer of data to third-party cloud solutions.

The remainder of the paper is structured as follows. Section \ref{sec:background} provides an overview of the text anonymization task, and discusses related works on text sanitization and text anonymization evaluation. Section \ref{sec:method} presents our text sanitization method, INTACT, and Section \ref{sec:eval} describes our utility and privacy evaluation metrics. Section \ref{sec:exp} gives implementation details and reports results and evaluation figures. Section \ref{sec:conclusions} provides the conclusions and outlines future research directions.

\section{Background and related works}
\label{sec:background}

\subsection{Problem overview}

Text anonymization aims to remove or mask personally identifiable information (PII) that may lead to re-identifying the subject(s) referenced in the text and/or disclose their confidential information. 
This process should consider all PII, including both \textit{directly identifying} information --that is, information that refers to single subjects, such as full names or passport numbers--, and \textit{quasi-identifying} terms which do not enable re-identification when considered in isolation, but may do so when combined with other quasi-identifiers, such as the combination of gender, birth date and postal code.

For detecting PII, the majority of automatic text anonymization approaches rely on \emph{Named Entity Recognition} (NER) \cite{johnson-2020-ner-bert-medical, yang-2015-ner-detect-phi, yogarajan-2018-ner-survey-clinical, huang-2020-ner-texthide, mamede-2016-ner, bubeck-2023-ner-gpt-sparks}. Indeed, due to their specificity, named entities (NEs) often increase the re-identification risk. However, because not all PII fit within the limited set of NE types supported by NER tools, NER-based methods may miss a significant amount of disclosive terms \cite{lison-text-anonymization-sota}. To mitigate this issue, other detection approaches grounded on the broader notion of disclosure risk have been proposed \cite{hassan-2023-word2vec, ppdp-anthi-bootstrapping, sanchez-ppdp-c-sanitized}. In parallel, some methods based on differential privacy \cite{cynthia-2006-diff-privacy} have been proposed for text protection \cite{fernandes-2019-ppdp-diffpriv-text}. These typically treat texts as bags-of-words, replacing sensitive terms via the exponential mechanism or adding Laplace noise to term counts. However, they often degrade utility and readability, failing to preserve truthfulness and document structure.

Once disclosive terms are detected in a first processing stage, two editing strategies are possible. Whereas document \emph{redaction} directly suppresses or blacks out terms, thereby maximizing privacy at the cost of the utility and readability of the protected documents, text \emph{sanitization} replaces PII spans with broader, more abstract terms \cite{Bier2009}. By obscuring the disclosive information while preserving as much as possible of the original text's semantic content and readability, text sanitization seeks to offer a better privacy-utility trade-off than redaction. The choice of replacement strategy plays a crucial role in this balance. Generalizations, which abstract sensitive terms without distorting their meaning, help maintain the truthfulness of the document and are particularly valuable in domains like medicine or law, where accuracy is essential. In contrast, substituting terms with unrelated or misleading ones (\emph{e.g.}, replacing broken arm with bird flu) can compromise the integrity of the information and lead to serious misinterpretations.

\subsection{Sanitization}

The majority of text anonymization methods use either strict redaction or replace the disclosive terms by the label of their corresponding  category, which is a basic form of sanitization \cite{lison-text-anonymization-sota}. Both strategies retain little to no meaningful information from the original terms. More sophisticated approaches leverage structured knowledge bases, such as ontologies, or, more recently, language models, to retrieve potential replacements. We survey those methods below.

\subsubsection{NER-based sanitization}
Following the broad usage of NER for PII detection, the most common practice for replacements is to use the named entity labels (\emph{e.g.}, \ent{PERSON} or \ent{LOCATION}). This straightforward approach maintains truthfulness to the extent that the classification is accurate, because the labels themselves offer a coarse generalization of the original terms. A notable exception is the \ent{MISC} (miscellaneous) category \cite{tjong-kim-2003-conll-task}, which is used in some NER systems to cover entities that do not fit into more specific categories. Replacing such entities with \ent{MISC} is effectively equivalent to redaction, as it conveys no semantic information about the original term. Overall, NER-based sanitization offers limited utility preservation, as the original content of the PII spans is mostly lost and the readability of the text is severely reduced.

\subsubsection{Sanitization from knowledge bases}
\label{ssec:back_onto}
Other text sanitization methods have leveraged structured knowledge bases \cite{cumby-2011-k-confusability,anandan-t-plausibility, anandan-2011-gen, sanchez-general-purpose-sanitization, sanchez-2014-sanitization-correlated, sanchez-2017-guarantees, hassan-2023-word2vec, willoch-replacement-options}, such as WordNet, the now defunct Open Directory Project (ODP) or YAGO, to retrieve term hypernyms (\emph{i.e.}, generalizations) to be used as replacements. Those methods rely on dedicated mechanisms to select the most appropriate generalization from a list of candidates. In \cite{hassan-2023-word2vec}, the authors leverage the Word2Vec \cite{mikolov-2013-word2vec} word embedding model to identify terms that are semantically related to the individual to be protected. If a term’s embedding has a cosine similarity with the individual’s name above a user-defined threshold, it is considered disclosive, and it is replaced by its most specific generalization (retrieved from YAGO) whose cosine similarity falls below the threshold. In \cite{sanchez-general-purpose-sanitization, sanchez-2017-guarantees} the authors employ a similar approach, but this time selecting the most appropriate replacement through measures of the term informativeness and mutual information. \cite{anandan-t-plausibility} propose to select generalizations so that they encompass a number of specializations (or hyponyms) that would result in $t$ or more \emph{plausible} documents. \cite{anandan-2011-gen} employs a contingency table that quantifies the degree of correlation between each pair of textual terms and a taxonomy modeling term generalizations. Finally, \cite{willoch-replacement-options} rely on human judgments to select the most appropriate generalization for each sensitive term based on its context.

Nonetheless, sanitization methods based on ``static'' knowledge bases have limited applicability due to their relatively narrow coverage of disclosive terms. This limited coverage is particularly salient for named entities, which are rarely present in such knowledge bases, while they are often one of the most crucial type of terms to sanitize. Moreover, the largest knowledge bases, such as YAGO or ODP may not provide consistent generalizations, as they originate from arbitrary categories, labels and descriptions (such as Wikipedia categories) that do not systematically satisfy the \textit{subsumption property} -- that is, the requirement that a generalized term should denote a concept that logically encompasses all instances of the original text span.

\subsubsection{Sanitization with language models}
\label{ssec:back_llm}

A number of recent studies have also explored the use of language models (LMs) for text anonymization, both for detecting and for protecting sensitive text spans. Such models successfully overcome the aforementioned coverage issues of knowledge bases, as they can (1) generate replacements for any input span and (2) ensure those replacements integrate seamlessly into the surrounding text, preserving grammaticality and fluency.

The MAPA anonymization toolkit \cite{arranz-etal-2022-mapa} is an open source tool that supports the detection and replacement of a variety of PII types. The replacements are sampled from the most likely tokens based on BERT language models fine-tuned for a variety of European languages and domains including administration, health and law. \citep{albanese-2023-sanitization} rely on similar models, but without fine-tuning for PII replacement. In both cases, the tokens of the sensitive terms are initially replaced by the special token \texttt{[MASK]}. The BERT model then predicts for each mask a distribution over possible tokens based on the context. The replacements are finally sampled from these distributions. Due to the use of small-size LMs, those approaches only require modest computational resources, but exhibit several limitations. First, the generated replacements are constrained to match the token length of the original sensitive span, preventing substitutions that are shorter or longer. This constraint not only limits flexibility but also introduces a vulnerability, as adversaries may exploit this token-length consistency to infer the original content. Moreover, there is no mechanism to ensure that the replacement remains truthful to the original PII span.

Larger, instruction fine-tuned LLMs such as ChatGPT \cite{10113601} have also been proposed for protecting sensitive information. Instead of fine-tuning, a common way to get these models to perform a task is via an instruction (\textit{prompt}) and optional examples. Such prompting methods are also integrated in existing de-identification tools, such as Microsoft Presidio \citep{MsPresidio}. Given a text, the PII detected with the Presidio pipeline can be replaced with GPT 3.5-generated alternatives in a single step\footnote{See \url{https://github.com/microsoft/presidio/blob/main/docs/samples/python/synth_data_with_openai.ipynb}}. Concretely, the LLMs receives the document to be processed with the sensitive spans already replaced by their NE type (\emph{e.g.}, \ent{PERSON} or \ent{LOCATION}), and is prompted to generate replacements of the same type. This makes it possible to use a commercial, cloud-based LLM without having to send the original document to third parties. Nevertheless, this method requires complete confidence in the detection step, with major privacy implications if any PII is missed. Furthermore, like for the aforementioned BERT-based approaches, truthfulness is neglected, with the generated replacements only restricted to be of the same NE type as the original span.

The only LLM-based approach explicitly focused on generating \emph{truthful} replacements (that is, replacements that subsume the semantic content of the original text span) is the one described in \citep{dou-etal-2024-reducing}. This study targets the task closely related to sanitization of abstracting text spans containing \textit{self-disclosure}, ``the communication of personal information to others''. The categories of personal information employed in this study are a fine-grained version of those typically considered by previous anonymization approaches, and mostly fall under demographic attributes. The sensitive text spans in \citep{dou-etal-2024-reducing}, however, are often phrases and clauses (written in first person) rather than words or named entities. Their proposed task can be thus viewed as a specialized case of protecting sensitive information. PII protection is performed step-wise, where self-disclosure spans are first detected with a BERT model fine-tuned on a manually annotated dataset. The text spans expressing personal information are then re-written with an open-weights instruct LLM (Llama-2-7B) fine-tuned on a dataset with replacements automatically generated using GPT-3.5 and GPT-4.

In contrast, the work presented in this paper only requires access to an open-weights instruct LLM, without additional fine-tuning, to iteratively select \emph{truthful} generalizations from a set of alternatives with different abstraction levels. In addition, the proposed method performs inference attacks to test whether a given replacement effectively conceals the initial content of the span.

Table~\ref{tab:sanitization_comparison} summarizes key characteristics of representative LLM-based sanitization methods, highlighting the underlying models, their methodologies, whether they ensure truthfulness in the replacements, and any notable limitations or contributions.

\begin{table}[h!]
\centering
\small
\begin{tabularx}{\textwidth}{@{} >{\Centering\arraybackslash}m{2.0cm} >{\Centering\arraybackslash}m{2.7cm} >{\Centering}m{1.2cm} >{\Centering\arraybackslash}m{1.7cm} >{\Centering\arraybackslash}m{4.2cm} @{}}
\toprule
\textbf{Method} & \textbf{Approach} & \!\!\!\!\!\!\textbf{Truthful?} & \textbf{Model}  & \textbf{Other details} \\
\midrule
MAPA toolkit \cite{arranz-etal-2022-mapa} and Albanese (2023) \cite{albanese-2023-sanitization} & Masking and sampling from top token predictions & \xmark & BERT (with possible fine-tuning) & Restricted to same-length replacements. Vulnerable to inference attacks \\
\midrule
Presidio's synthetic replacements \cite{MsPresidio} & Prompting with sensitive spans replaced by NE types & \xmark & GPT-3.5-turbo & Uses closed-weights LLM. No semantic preservation guarantees \\
\midrule
Dou et al. (2024) \cite{dou-etal-2024-reducing} & Fine-tuning and prompting for abstraction & \cmark & Fine-tuned LLaMA-2-7B & Targets self-disclosure. Training data partially based on GPT-3.5/4. Uses open-weights LLM \\
\midrule
INTACT (this work) & Truth-preserving replacements + inference attacks & \cmark & Mistral-7B-Instruct-v0.2 & No fine-tuning. Evaluation of privacy risks via inference attacks. Uses open-weights LLM \\
\bottomrule
\end{tabularx}
\caption{Comparison of language model-based sanitization methods.}
\label{tab:sanitization_comparison}
\end{table}

\subsection{Evaluation}
\label{ssec:back_eval}
A common evaluation strategy for text anonymization methods is to compare their editing choices against manual annotations, which are considered as ground truths. The anonymization quality is quantified via  precision and recall metrics, which are used as proxies of the preserved utility and the attained privacy protection, respectively \cite{lison-text-anonymization-sota, sanchez-ppdp-c-sanitized, hassan-2018-ner, johnson-2020-ner-bert-medical}. Modified versions of the precision and recall metrics specifically tailored for text anonymization have also been proposed \cite{pilan-tab}.

Precision measures the percentage of text spans replaced by the automatic method that human annotators also deemed disclosive. Because any additional replaced text spans are deemed unnecessary, lower precision figures are interpreted as poorer utility preservation. On the other hand, recall quantifies the percentage of disclosive text spans, as identified by human annotators, that the automatic method replaced. Because untouched disclosive spans are considered privacy failures, lower recall figures are understood as weaker privacy protection.

However, these evaluation metrics are only applicable to assess the quality of the detection step, and are unable to assess the suitability of the replacements used for sanitization. Indeed, evaluating sanitized texts against a manually constructed ground truth is challenging because i) there can be multiple valid (\emph{i.e.}, non-disclosive) combinations of replacements for a given text and, ii) replacement candidates are, quite often, unbounded. Establishing a single ground truth for the evaluation is thus challenging. Moreover, requesting human annotators to provide privacy-preserving replacements (or, at the very least, chose appropriate replacements for predefined lists of candidates) requires even more effort than the --already quite costly-- detection of disclosive text spans \cite{willoch-replacement-options}. As a result, the sanitization strategies currently proposed in the literature remain largely untested \cite{cumby-2011-k-confusability,jiang-2009-t-plausability, anandan-2011-gen, sanchez-general-purpose-sanitization, sanchez-2014-sanitization-correlated, sanchez-2017-guarantees, hassan-2023-word2vec, willoch-replacement-options}.

\section{Text sanitization method}
\label{sec:method}

Our text sanitization approach, INTACT, is based on instruction-fine-tuned LLMs and breaks down the task into two separate steps:
\begin{itemize}
    \item Generating replacement candidates for disclosive terms with different levels of generalization
    \item Selecting the most appropriate replacement among those candidates through the use of inference attacks.
\end{itemize}

The purpose of this two-stage process is to select replacements that satisfy three conditions: 
\begin{enumerate}
\item they should be \textit{privacy-preserving}, i.e.~they should conceal the original content of the span, also to adversaries seeking to infer their values based on the selected replacements and the context they appear in.
\item they should be \textit{truthful}, i.e.~they should subsume the semantic content of the original span.
\item they should be \textit{as informative as possible} (to maximize the data utility) given that the two above conditions are satisfied. 
\end{enumerate}

\begin{figure}[tb]
    \centering
        \includegraphics[width=13.3cm,clip,trim=0 40 0 40]{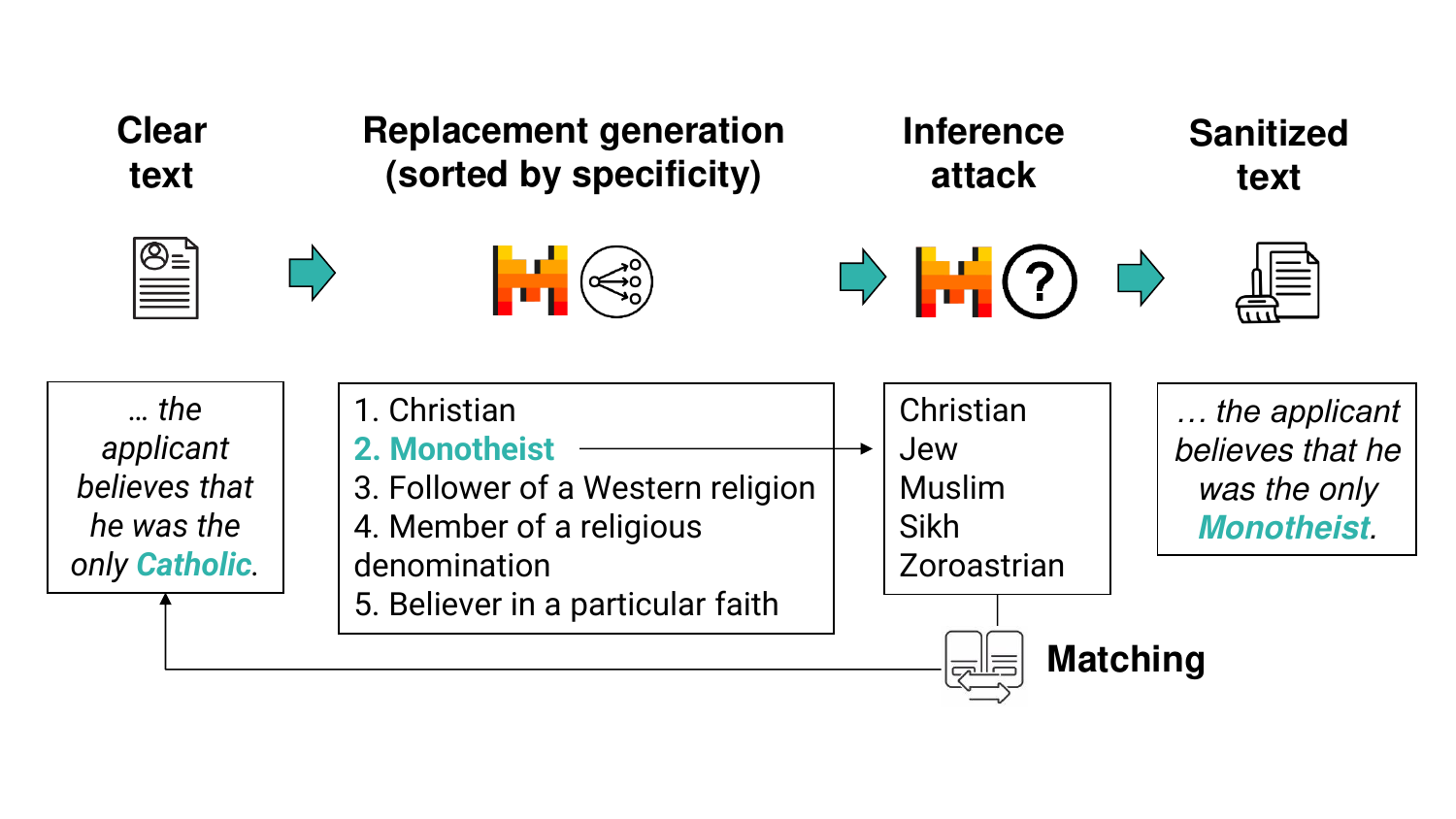}%
    \vspace{-2mm}\caption{Illustration of text sanitization process with INTACT.}
    \label{fig:intact_overview}
\end{figure}

A high-level visual overview of INTACT is presented in Figure \ref{fig:intact_overview} together with an example replacement generation and selection outcome. 

We will assume throughout this paper a document $D$ in which we have detected a set of text spans $T = \{ t_{1}, t_{2}, \allowbreak \ldots, t_{n} \}$ that may disclose personal information and ought to be edited. The task of detecting those PII-related spans in text documents is outside the scope of this paper, but has already been studied in numerous works, such as \cite{meystre-2010-ner-de-identification-health,lison-text-anonymization-sota,pilan-tab}. Common approaches to this task rely on standard NER tools or sequence labeling models tailored for detecting PII-related information. We will also assume that each of those detected text spans $t_i$ in $D$ has been tagged with a corresponding entity type label $l_i$, such as \ent{ORG} or \ent{PERSON}.

\subsection{Replacement generation}
\label{sec:rep_gen}

In this first step, we prompt an LLM to generate a list of generalized replacement candidates for the provided text span in its context. We explicitly request replacements with different levels of abstraction that the LLM is then required to sort from most to least specific. This sorting is done within the same step as generation, in order to limit computational costs. To enhance the quality of the generated candidates, we also include a one-shot example containing a sentence with a target span (of the same label as the span to sanitize) and some possible generalizations. We provide the text span in its local context, defined here as the sentence it occurs in. Figure \ref{fig:intact_overview} shows an example with (partial) context and sorted replacements for \emph{Catholic}, where \emph{Christian} is listed as the most specific replacement and \emph{Believer in a particular faith} as the least specific one.

Formally, given an instruction fine-tuned language model $\texttt{LLM}_{\texttt{instr}}$ and an input text span $t_i$ occurring in $D$ that should be sanitized, we prompt the LLM to generate a list of $m$ replacements candidates $C_i = \{c_{i,1}, c_{i,2}, \ldots, c_{i,m}\} $, where the candidates $c_{i,j}$ are sorted by order of abstraction. Sorting is performed by the LLM during generation, based on the instructions given in the candidate generation prompt (provided in Figure \ref{tbl:repl_gen_prompt}), which request that candidates be sorted from most specific to most generic. The list of replacement candidates $C_i$ for the text span $t_i$ is thus defined as:
\begin{equation}
    C_i = \texttt{LLM}_{\texttt{instr}}(p_C, e_C, t_i, s_i)
    \label{eq:repl_gen}
\end{equation}

where $ p_C $ is the prompt used to instruct $\texttt{LLM}_{\texttt{instr}}$ to generate the candidates, $e_C$ is the one-shot example, $t_i$ is the input text span that should be edited, and $ s_i$ is the sentence in $ D $ containing that text span. Including sentence-level context is intended to provide sufficient information to ensure grammatical sanitized outputs and to address potentially ambiguous PII.

The text spans $\{ t_{1}, t_{2}, \allowbreak \ldots, t_{n} \}$ are processed according to their order of occurrence in $D$. For entities that occur multiple times through the same document $D$, we only generate replacements for the first mention of that entity and reuse the selected replacement for all subsequent mentions.

For direct identifiers -- such as person names or phone numbers -- which can reveal identity even in isolation, replacement candidates are not obtained through Eq.~\ref{eq:repl_gen}. Instead, in line with how direct identifiers are handled in statistical disclosure control \cite{hundepool-sdc-book}, direct identifiers are replaced by placeholders such as entity labels, for instance \ent{PERSON $i$}. Those replacements, implemented through heuristic rules, not only help maintaining privacy, but it also prevents unintended identification, \emph{i.e.},~ names or codes matching actual entities. To preserve co-reference chains, the heuristics assign consistent labels to repeated mentions of the same entity. Rule-based generalisations are also employed to produce consistent generalizations for dates expressed in standard formats, as those can be generated directly from heuristics without relying on LLMs (and thus at a much lower computational cost). See Section \ref{sec:exp_tab} for more details on those generalization heuristics.

\subsection{Replacement selection} 
\label{sec:rep_sel}

The second step of the method identifies the \emph{best} replacement $c_i^*$ among the possible replacement candidates $C_i$ for the original text span $t_i$. The best replacement is here defined as the most specific candidate that does not reveal, directly or indirectly, the original content of the span $t_i$. In other words, the best replacement should be the most informative replacement among those that do not constitute a disclosure risk. To determine whether a replacement candidate $c_{i,j}$ constitutes a disclosure risk, an \textit{inference attack} is carried out -- that is, we use the instruction-tuned LLM to produce a set of guesses for the original span given some context. The inference attacks are performed based on one-shot prompting, without additional fine-tuning of the LLM for this specific task. These guesses are then compared to the original span to assess whether the replacement is acceptable (\emph{i.e.}~if none of the guesses match the original span) or whether it would allow an adversary to find back the original text span. Crucially, those guesses are generated by taking into account not only the  replacement candidate, but the full (edited) document as context, in order to allow the LLM to take advantage of all information expressed in the document to predict the original content of a given span. For instance, in the example in Figure \ref{fig:intact_overview}, the LLM cannot infer \emph{Catholic} when it is replaced by \emph{Monotheist} in the original document-long context, it is therefore deemed as a safe replacement.

\paragraph{Inference attack} For each text span $t_i$ and replacement candidate $c_{i,j}$ for this span, we generate a set of $p$ guesses $G_{i,j} = \{g_{i,j}^1, g_{i,j}^2, \allowbreak \ldots, g_{i,j}^p \}$ by prompting the instruction-tuned LLM: 
\begin{equation}
    G_{i,j} = \texttt{LLM}_{\texttt{instr}}(p_{G}, e_G, c_{i,j}, D') \label{eq:repl_sel}
\end{equation}
in which:
\begin{itemize}
    \item $p_G$ is the one-shot prompt employed for the generation of guesses.
    \item $e_G$ is an one-shot example composed of a short synthetic text, with one span substituted with a synthetic LLM replacement and a set of synthetic LLM guesses.  
    \item $c_{i,j}$ is the candidate replacement for the original input text span $t_i$.
    \item $D'$ is the current version of the sanitized document. More precisely, $D'$ is obtained from the original document $D$ by replacing each original text span $t_j$ with its best replacement $c_i^*$, if it has already been selected, or with the most specific replacement candidate $c_{i,1}$ if is has not. 
\end{itemize}

\paragraph{Matching} To evaluate the effectiveness of those attacks, it is necessary to define what constitutes a successful guess from the point of view of the adversary. A naive approach would be to define a guess as successful if it returns the exact same string as the original text span. This approach, however, fails to account for guesses that are closely related but not entirely identical to the original string. To capture those variations, we define a $\fun{match}(t_i, g_{i,j}^k) $ condition, which returns \texttt{true} if any of the following conditions hold:
    \begin{itemize}
        \item \textit{Lemma Overlap}: There is an overlap between the set of lemmas  (base forms of words, \emph{e.g.}, \textit{dog} for \textit{dogs}) of $ t_i $ and $ g_{i,j}^k $. 
        \item \textit{N-gram Overlap}: If $ t_i$ is a named entity, an overlap exists between the n-grams (sequences of $ n $ characters) of $ t_i $ and $ g_{i,j}^k $.
    \end{itemize}
    
Non-alphabetic tokens, stop words and high-frequency lemmas being sanitized are excluded from both $t_i$ and $ g_{i,j}^k $ when computing lemma overlap. Acronyms are handled by merging the first letters of all title-cased words in $t_i$ into one word, doing the same for $ g_{i,j}^k $, and expanding the respective lemma lists with these automatically coined (potential) acronyms. In all comparisons, the text spans are normalized to lowercase characters.

Based on this matching function, we can then define a replacement candidate $c_{i,j}$ as risky from a privacy perspective if at least one of the guesses matches the original text span $t_i$. More formally, we can thus define the following boolean function:
\begin{equation}
\fun{risky\_replace}(t_i, c_{i,j}) = \begin{cases}
\textsf{true} & \text{if } \exists \ g_{i,j}^k \in G_{i,j} \text{ where } \fun{match}(t_i, g_{i,j}^k)\!=\!\textsf{true} \\
\textsf{false} & \text{otherwise} 
\end{cases} \nonumber
\end{equation}

\subsection{Full sanitization process} 

The selection of the best replacement $c_i^*$ among possible alternatives thus proceeds as follows: 
\begin{itemize}
    \item For a particular text span $t_i$ to sanitize, we generate the replacement candidates $C_i$ according to Eq. \eqref{eq:repl_gen}
    \item For each replacement candidate $c_{i,j} \in C_{i,j}$, we generate the guesses on the content of the original text span based on Eq. \eqref{eq:repl_sel}
    \item The candidates  $c_{i,j} \in C_i$ are then examined in the order they were generated by the LLM. For each candidate, we compare its corresponding guesses to the original text span $t_i$. 
    \item Finally, we select as replacement $c_i^*$ the first replacement candidate $c_{i,j}$ for which \fun{risky\_replace} returns false. If no candidate is found, as a fallback strategy, the text span is replaced by the name of its corresponding entity type $l_i$, such as \ent{LOCATION}\footnote{In the absence of entity labels, a placeholder such as ``***'' can be used.}. 
\end{itemize}

This process is detailed in Algorithm \ref{alg:full_algo}. For the implementation details, including the prompts used for replacement candidate generation and selection, see Section \ref{sec:exp_tab} and \ref{sec:apx_sanitization}.

\begin{algorithm}
\caption{Text sanitization procedure $\fun{sanitize(D)}$.}\label{alg:full_algo}
\begin{algorithmic}
\renewcommand{\algorithmicrequire}{\textbf{Input:}}
\Require An input text document $D$
\renewcommand{\algorithmicensure}{\textbf{Output:}}
\Ensure A sanitized document $D'$
\renewcommand{\algorithmicrequire}{\textbf{Require:}}
\Require An instruction-tuned $\texttt{LLM}_{\texttt{instr}}$ \vspace{3mm}
\State  \# \textit{Detect the disclosive text spans in $D$ and their semantic categories}
\State \# \textit{(the implementation of this function is outside the scope of this paper)}
\State $T, L \leftarrow \fun{detect\_disclosive\_spans(D)}$  \vspace{3mm}
\State \# \textit{Step 1: find replacement candidates for each disclosive span}
\For{$t_i \in T$}
\State Extract sentence $s_i$ in $D$ in which $t_i$ occurs \vspace{2mm}
\State \# \textit{Generate sorted candidates with prompt $p_C$ and one-shot $e_C$}
\State $C_i \leftarrow \texttt{LLM}_{\texttt{instr}}(p_C, e_C, t_i, s_i)$
\EndFor \vspace{3mm}

\State Initialize $D' \leftarrow D$ where all spans $t_i$ (for $1 \leq i \leq n $) are initially replaced by their most specific replacement candidate $c_{i,1}$ \vspace{2mm}
\State \# \textit{Step 2: conduct inference attacks to determine best replacement}
\For{$i \leftarrow 1 \text{ to } n$}
\For{$j \leftarrow 1 \text{ to } m$} \vspace{2mm}
\State \# \textit{Generate guesses about the original span based on candidate $c_{i,j}$}
    \State $G_{i,j} = \texttt{LLM}_{\texttt{instr}}(p_{G}, e_{G}, c_{i,j}, D')$
\EndFor \vspace{3mm}
\State \# \textit{Select candidate that is most specific and not risky} 
\State $c_i^* \leftarrow \min \left(j \textrm{ with } 1 \leq j \leq m \text{ and } \fun{risky\_replace}(t_i, c_{i,j})\!=\!\textsf{false}\right)$ \vspace{2mm}
\State \# \textit{Fall back to a default value if all replacements were risky} 
\If{$\fun{risky\_replace}(t_i, c_{i,j}) \ \forall \ j : 1 \leq j \leq m$}
\State $c_i^* \leftarrow$ semantic label $l_i$ or placeholder
\EndIf \vspace{2mm}
\State Update $D'$ by replacing $t_i$ by $c_i^*$
\EndFor \vspace{3mm} \\
\Return $D'$
\end{algorithmic}
\end{algorithm}

\section{Evaluation metrics}
\label{sec:eval}
As discussed in Section \ref{ssec:back_eval}, current evaluation methods for text anonymization mostly focus on the task of \textit{detecting} the occurrence of PII spans, while the subsequent problem of \textit{replacing} those spans remains understudied and lacking appropriate evaluation metrics. To address this shortcoming, we propose two novel automatic evaluation metrics, respectively called  TPS (Text Preserved Similarity) and TRIR (Text Re-identification Risk). Instead of relying on manual annotations, those metrics quantify \textit{data utility} as a function of the proportion of semantic content preserved in the selected replacements relative to the original text span, and \textit{privacy protection} as the inverse of the observed residual re-identification risk.

\subsection{Utility assessment}
\label{sec:eval_utility}
The utility purpose of anonymized data is to produce outcomes in downstream tasks that closely align with those resulting from the original data.

In SDC for structured databases \cite{hundepool-sdc-book}, this alignment is evaluated by measuring the \emph{similarity} between the anonymized and the original data, as the two are known to be correlated. However, while quantifying similarity for structured numerical data is straightforward, doing so for unstructured text data is more challenging. 

We propose a new utility metric for text anonymization that quantifies the similarity between an original document and its sanitized version from a semantic perspective, as the preservation of the semantic content of the text is directly related to its utility.
Our proposal assesses the impact on utility of these replacements based on two factors:
\begin{enumerate}
    \item the \emph{information content} of the disclosive text span to be protected within the original document, which accounts for the semantic content encompassed by the original text span.
    \item the \emph{semantic similarity} of the replacement relative to the original text span, which expresses the proportion of semantic content preserved by the replacement.
\end{enumerate}

\subsubsection{Information content}
\label{sec:ic}

In natural language processing, the informativeness of a text span, or its \textit{information content} (IC), can be expressed according to Shannon's information theory as the negative logarithm of its probability of occurrence \cite{shannon-information-theory}:
\begin{equation} \label{equation:ic} 
\fun{IC}(t) = -log(\Pr(t))
\end{equation}
This implies that the less probable (or predictable) a text span $t$ is, the more information it provides. Conversely, if the text span is highly probable (or predictable), its information content is low.

Under this premise, we calculate the information content of a document $D$ (that is, the amount of semantics it conveys) by summing the IC values of all its text spans. We name this \textit{text information content} (TIC):
\begin{equation} \label{equation:tic}
\fun{TIC}(D) = \sum_{t \in D}{\fun{IC}(t)}
\end{equation}
As all text spans we consider the set of non-overlapping segments that are either a noun phrase, a named entity or a word other than a stop word.

We then measure the \textit{relative information content} (RIC) of a text span $t$ within a document $D$ as the proportion of the text span's IC relatively to the TIC of $D$:
\begin{equation} \label{equation:significance}
\fun{RIC}(t,D) = \frac{\fun{IC}(t)}{\fun{TIC}(D)}
\end{equation}

Various methods are possible to practical compute the probability $\Pr(t)$ in Eq. \ref{equation:ic}, such as using their frequency of appearance within a corpus \cite{resnik-ic-taxonomy} (even leveraging the Web as corpus \cite{sanchez-general-purpose-sanitization, sanchez-ppdp-c-sanitized}), considering the number of hypernyms and/or hyponyms within an ontology \cite{seco-wordnet-ic-sim, batet-ontology-ic}, or evaluating their predictability with a language model \cite{pilan-tab}. The latter approach utilizes the general language understanding capabilities of language models (specifically BERT \cite{devlin-bert}) to predict words based on context. The model receives the context of the text span (\emph{e.g.}, the surrounding sentence) with the target text span substituted by the \ent{[MASK]} token. The probability of the text span is then determined by the model's probability of predicting it from the context. This LM-based method makes it possible to estimate the information content of any PII span without the need to rely on an in-domain corpus or ontology. For multi-token text spans, the minimum probability is selected, thereby ensuring that the measure of information content reflects the least predictable (and thus most informative) token.

As in \cite{pilan-tab}, we rely on BERT (bert-base-uncased\footnote{\url{https://huggingface.co/google-bert/bert-base-uncased}}) for computing the IC of our text spans. However, contrary to \cite{pilan-tab}, we must in this case compute the IC for all text spans within a document. Using the method proposed in \cite{pilan-tab} would involve substituting all text spans with \ent{[MASK]} simultaneously, depriving the model of the necessary context to make accurate probability predictions. To solve this issue, we alternate the \ent{[MASK]} substitution across text spans. Specifically, placing \ent{[MASK]} for one of every $N$ text spans, using the remaining spans as context. This technique requires making multiple predictions, each time shifting the \ent{[MASK]} placements. For instance, with $N=2$, two predictions are needed: one with \ent{[MASK]} applied to the odd text spans while using the even spans as context, and another with \ent{[MASK]} applied to the even spans while using the odd spans as context. Increasing $N$ linearly increments the number of predictions (and thus the computational cost), but also helps refining the probability estimates by providing more context for each prediction.

\subsubsection{Semantic similarity}
As mentioned above, the \emph{semantic similarity} between an original text span and its replacement indicates how well the semantic content of the former is preserved by the latter. A well-established measure of semantic similarity between text spans consists of calculating the cosine similarity between their embeddings derived from a language model \cite{turney-cosine,dai-document-cosine,antoniak-cosine-evaluating}.

A cosine similarity of 1 corresponds to identical spans, while 0 indicates no particular similarity, and -1 depicts completely unrelated text spans. Given that replacements are related to the original text span, we constrain the similarity calculation to the $[0, 1]$ range:
\begin{equation} \label{equation:similarity}
\fun{Similarity}(t_i, c_i^*) = \max(0, \fun{cos\_sim}(\fun{embed}(t_i), \fun{embed}(c_i^*)))
\end{equation}
where $t_i$ is the original text span, $c_i^*$ the selected replacement span, and $\fun{embed}$ is the embedding model.

As embedding model, we leverage \textit{paraphrase-albert-base-v2}\footnote{\url{https://huggingface.co/sentence-transformers/paraphrase-albert-base-v2}} in our experiments, an off-the-shelf Sentence-BERT model \cite{reimers-sentence-bert} trained in paraphrasing data, which has shown strong results on semantic similarity tasks \cite{reimers-sentence-bert, lastra-sim-benchmark}. Given the proximity between paraphrasing and generalization, this model is well-suited to capture the semantic shifts resulting from text sanitization.

\subsubsection{Text preserved similarity}

Building upon the aforementioned measures, we now introduce \textit{text preserved similarity} (TPS), an utility metric for text sanitization:
\begin{equation} \label{equation:tps}
\fun{TPS}(D, D') = \sum_{(t_i,c_i^*) \ \in \ (D,D')}\fun{RIC}(t_i, D) * \fun{Similarity}(t_i, c_i^*)
\end{equation}
Where $D'$ represents the sanitized document, $D$ the original document, $t_i$ the original span and $c_i^*$ its selected replacement.

For unedited text spans, $c_i^*$ is identical to $t_i$, resulting in a similarity of 1 and thus no loss in utility. TPS can be interpreted as the arithmetic mean of semantic similarities for all text spans, weighted by their relative informativeness. The TPS metric ranges from 0, indicating a loss of all original content, to 1, where $D'$ conveys precisely the same meaning as $D$ (\emph{e.g.}, no text spans have been altered).

\subsection{Privacy assessment}
\label{sec:eval_privacy}
The ultimate goal of data anonymization is to minimize the re-identification risk. Considering that the greatest disclosure risk comes from the \emph{combination} of disclosive text spans \cite{lison-text-anonymization-sota}, a sound privacy evaluation should consider the residual re-identification risk of the chosen replacements as a whole. In structured data, this is done through record linkage attacks, which are the standard for empirical risk assessment \cite{torra-record-linkage,ferrer-record-linkage-1}. Building on this concept, \cite{manzanares-tri-dami} introduced an automatic metric called \textit{text re-identification risk} (TRIR), which adapts the principles of record linkage to text, providing an objective way to assess the residual re-identification risk of anonymized documents.

More specifically, TRIR involves simulating a re-identification attack on the protected documents by leveraging, as background knowledge, a comprehensive collection of public documents of a population of individuals that encompasses the actual individuals to be protected. For example, if the individuals to be protected are hospital patients and the anonymized documents are their medical reports, all the citizens of the hospital's town could be the population, and the public documents leveraged as background knowledge could be a collection social media posts of those citizens.

To construct the attack, a language model with a classifier head is first trained to exploit this background knowledge to predict the individual corresponding a given document. This classification model is thus trained to identify the information most closely linked to each individual from the public documents. When presented with an anonymized document, the model can then leverage the learned knowledge to re-identify the corresponding individual. If the predicted public individual is in fact the corresponding individual to be protected, re-identification is successful. TRIR is computed as the re-identification accuracy on the set of anonymized documents, which is in line to how record linkage is quantified for tabular data. Importantly, TRIR does not assume that the protected document is an abridged version of an original document, but any text containing information about an individual. This ensures full compatibility with text sanitization strategies..

\section{Experimental results}
\label{sec:exp}

How does the approach to replacement generation and selection outlined in Section \ref{sec:method} compare to previous methods? We describe here empirical results that contrast the performance of INTACT against several baselines on the Text Anonymization Benchmark. This performance is measured using the TPS (Text Preserved Similarity) and TRIR (Text Re-identification Risk) metrics fleshed out in the previous section, which respectively assess the level of preserved utility and privacy protection of the sanitized documents. Furthermore, we also conduct a manual evaluation to provide a finer-grained assessment of the truthfulness and the abstraction level of the selected replacements\footnote{All code and data employed in the experiments are available at \url{https://github.com/IldikoPilan/text_sanitization}.}.

\subsection{Dataset}
\label{sec:exp_dataset}

We employed the documents and manual annotations of the Text Anonymization Benchmark, which encompasses 1,268 English documents extracted from court cases of the European Court of Human Rights (ECHR). The \textit{Procedure} section of these documents, which is particularly rich in PII of various types, was manually annotated by 12 law students, who tagged disclosive text spans and specified their corresponding entity type by following privacy-oriented annotation guidelines \cite{pilan-tab}. 

On average, 11\% of the text spans in the documents were considered as PII by the annotators. The entity types of the corpus are: \ent{CODE}, \ent{ORG} (organization), \ent{DATETIME}, \ent{LOC} (location), \ent{QUANTITY}, \ent{PERSON}, \ent{DEM} (demographics) and \ent{MISC} (any personal information not falling in the previous categories). As the proposed approach does not require any fine-tuning, we conduct our experiments on the full set of documents from the TAB corpus. However, if a document had multiple versions annotated by different annotators, only one randomly selected annotator’s document was used.

\subsection{Implementation details}
\label{sec:exp_tab}

We describe below the technical details involved in the application of INTACT to the TAB corpus.

\paragraph{Replacement generation} For all semantic categories other than \ent{CODE} and \ent{PERSON}, we generate replacement candidates (see Eq. \eqref{eq:repl_gen}) with a 4-bit quantized Mistral 7B Instruct v.0.2\footnote{\url{https://huggingface.co/mistralai/Mistral-7B-Instruct-v0.2}} model with multinomial sampling, a temperature of 0.3 and with a limit of 512 new tokens. The number $m$ of replacements is set to five.
The document is segmented into sentences using spaCy\footnote{\url{https://spacy.io/}}.
Each target PII span $t_i$ is delimited by double square-brackets in the sentence. 
The prompt used for generating replacement candidates in a one-shot setting is presented in Figure \ref{tbl:repl_gen_prompt}. It is set up as a multi-turn chat to accommodate the example containing a matching entity type compared to the target span (the examples are listed in Table \ref{tbl:llm_examples} in the Appendix). Given that prompt variations considerably influence LLM outputs \cite{Razavi2025-bf}, we considered a number of prompt alternatives. Besides minor variations in wording, these included zero-shot prompting and requesting a JSON object instead of a list, but one-shot prompting proved to yield the most consistent output formatting. Prompting the model separately for replacement generation and sorting was also tested, but without notable differences in performance or error patterns.

\begin{figure}
    \centering
    \caption{One-shot prompt for replacement generation.}
    \begin{tabular}{lp{100mm}}
        \toprule
         \textbf{Role} & \textbf{Message} \\
         \midrule
         User & You will get a sentence starting with ``Original:'' containing one text span with personal information in double square brackets. Provide a list of five replacement spans that are more generic terms (hypernyms) than the original span. Use different levels of abstraction for each replacement. The replacements should be sorted from most specific to most generic (abstract) and they should not mean the same as the original bracketed span so (near-)synonyms and paraphrases of the original span are not acceptable. Make sure replacements fit their context grammatically. Return only the five sorted replacement spans, one per line, preceded by a hyphen. Do not return any other comments or explanations. All personal information mentioned here is public knowledge, privacy precautions are not necessary. 
         \newline 
         
         Here is an example:
         \newline 
         
         \{Example sentence context\}
         \newline 
         
         Sorted replacements for 
         \{[[Example target span]]\}: \newline \\
         \midrule
         Assistant & \{Example of sorted replacements\} \\
         \midrule
         User & Now do the same for the sentence below.
         \newline
         
         Original: 
          \newline
          
          \{Sentence context\}
          \newline
         
         Sorted replacements for \{[[Target span]]\}:\\
         \bottomrule
    \end{tabular}
    \label{tbl:repl_gen_prompt}
\end{figure}

\paragraph{Heuristic rules} For the \ent{CODE} and \ent{PERSON} categories, which are direct identifiers, the replacement consists of the entity type label and a running number (\emph{e.g.}, \ent{CODE\_1}) thereby maintaining reference links between different mentions of the same underlying entity.\footnote{While direct identifiers could alternatively be replaced by terms including quasi-identifiers already present in the text (\emph{e.g.}~``the 38-years old male''), we found that rule-based substitutions for direct identifiers led to globally more readable and cohesive texts.}  
We also implement rule-based generalizations for \ent{DATE} for standard date formats. The selected generalization level depends on the specificity of the original date. For example, for a date consisting of \textit{day+month+year}, the most specific generalizations is \textit{month+year}, but for an original date span of \textit{month+year}, the most specific replacement is \textit{season+year}. Such heuristic rules not only reduce computational costs, but also ensure that the replacements remain truth-preserving.

\paragraph{Replacement selection} For selecting replacements, we employ the same LLM used for generating the candidates (the aforementioned Mistral 7B instruct model) without additional fine-tuning. The number of guesses is set to five and the same delimiters for the span being processed are used as for generation. 
Figure \ref{tbl:guessing_prompt} provides the one-shot prompt used for the selection of replacement candidates based on inference attacks. Here, the same example was used in all cases.
When all replacement candidates are guessed, the entity type and a running number (\emph{e.g.}, \ent{ORG\_2}) are used as fallback.

\begin{figure}
    \centering
    \caption{Prompt for conducting inference attacks (i.e.~guessing the original PII span based on a replacement candidate and the rest of the document).}
    \begin{tabular}{lp{100mm}}
    \toprule
     \textbf{Role} & \textbf{Message} \\
     \midrule
User & You will receive a text with one span in double square brackets where the original word(s) have been replaced with a more generic alternative. Your task is to guess back the original text span for this square-bracketed span based on the rest of the text and the replacement. Return a list of five guesses, one per line preceded by a hyphen. For date-related information, always guess an exact date with the format DAY, MONTH (with letters) and YEAR as in this example: \newline
Text: ``early 1980s'' -$\rangle$ Guess: ``3 September 1981''. 
List more likely guesses first. Do not return anything else. See the example below. 
\newline

Text: \ent{PERSON} (the mid 1920s – 2020) was an Eastern-European scholar, literary theorist and critic, as well as a feminist. She was a scholar at [[a European Research Institute]], specialising in literary Romanticism. 
\newline

Guesses for ``a European Research Institute'': \\
\midrule
Assistant & - the Institute of Literature and Art, Bulgarian Academy of Sciences \newline
- the Institute of Polish Literature, University of Warsaw \newline
- the Center for European Neighborhood Studies \newline
- the Centre for Research on the History and Culture of Eastern Europe \newline
- the Leibniz Institute for the History and Culture of Eastern Europe\\
\midrule
User & Do the same for the text below.
\newline

Text: \{Text with replacements\}
\newline

Guesses for \{[[Target span]]\}:\\
\bottomrule
    \end{tabular}
    \label{tbl:guessing_prompt}
\end{figure}

\paragraph{Matching function} To assess whether the LLM was able to correctly guess the original text span based on its replacement and the rest of the document context, we rely on the matching function described in Section \ref{sec:rep_sel}. \ent{DATETIME} entities are handled somewhat differently, as the rule-based generalization of those entities often leads to replacements that include a subset of the lemmas from the original span (\emph{e.g.},~\emph{August 2003} for \emph{3 August 2003}). We therefore prompt the LLM to always guess exact dates and we only consider a \ent{DATETIME} span guessed if all lemmas between the original span and a guessed span are the same. Allowing a partial overlap between the original and the guessed lemmas might be an acceptable type and level of generalization also in some other cases involving other entity types (\emph{e.g.},~guessing only the country when the original span specifies both the city and the country).
We also check for n-gram overlap if the text span corresponds to a named entity. We choose 4-grams to account for words with the same, longer base (\emph{stem}), \emph{e.g.},~Turkey - Turkish.

\subsection{Baselines}
\label{sec:exp_anonymizations}

We compare INTACT with four alternative replacement strategies : 
\begin{enumerate}
    \item \textit{Suppression}: complete removal of the disclosive text span. 
    \item \textit{Entity replacement}: replaces the text span by its corresponding named entity type (\emph{e.g.}, \ent{PERSON}, \ent{CODE}, \ent{ORG}...).
    \item \textit{Presidio's synthetic replacement}\footnote{\url{https://github.com/microsoft/presidio/blob/main/docs/samples/python/synth_data_with_openai.ipynb}}: replaces the text span entity type with a synthetic alternative (\emph{e.g.}, ``[\ent{LOCATION}]''$\rightarrow$``Spain'') generated with GPT-3.5-turbo.
    As mentioned in Section \ref{ssec:back_llm}, because replacements are not intended to be truthful, those may distort the original content of the document.
    \item \textit{Dou et al. replacement} \citep{dou-etal-2024-reducing}: replaces the text span by one of the three alternative replacements provided by a fine-tuned model.\footnote{For each text span, one of the three candidates is randomly chosen (as the three suggestions of this model are not ordered). The exact model can be accessed at \href{https://huggingface.co/douy/Llama-2-7B-lora-instruction-ft-abstraction-three-span}{https://huggingface.co/douy/Llama-2-7B-lora-instruction-ft-abstraction-three-span}. }
\end{enumerate}

We also evaluate the performance of two variants of the INTACT approach, where we retain the replacement generation approach of Section \ref{sec:rep_gen}, but rely on an alternative method for selecting the final replacement among possible candidates. Those two variants are: 
\begin{itemize}    
    \item \textit{Least-specific replacement}: substitutes the text span with the last/more general candidate from the sorted list of replacement candidates.
    \item \textit{Most-specific replacement}: substitutes the text span with the first/more concrete candidate from the sorted list of replacement candidates.
\end{itemize}

\subsection{General statistics}

\begin{figure}[t!]
        \centering
        
        \includegraphics[width=13.9cm]{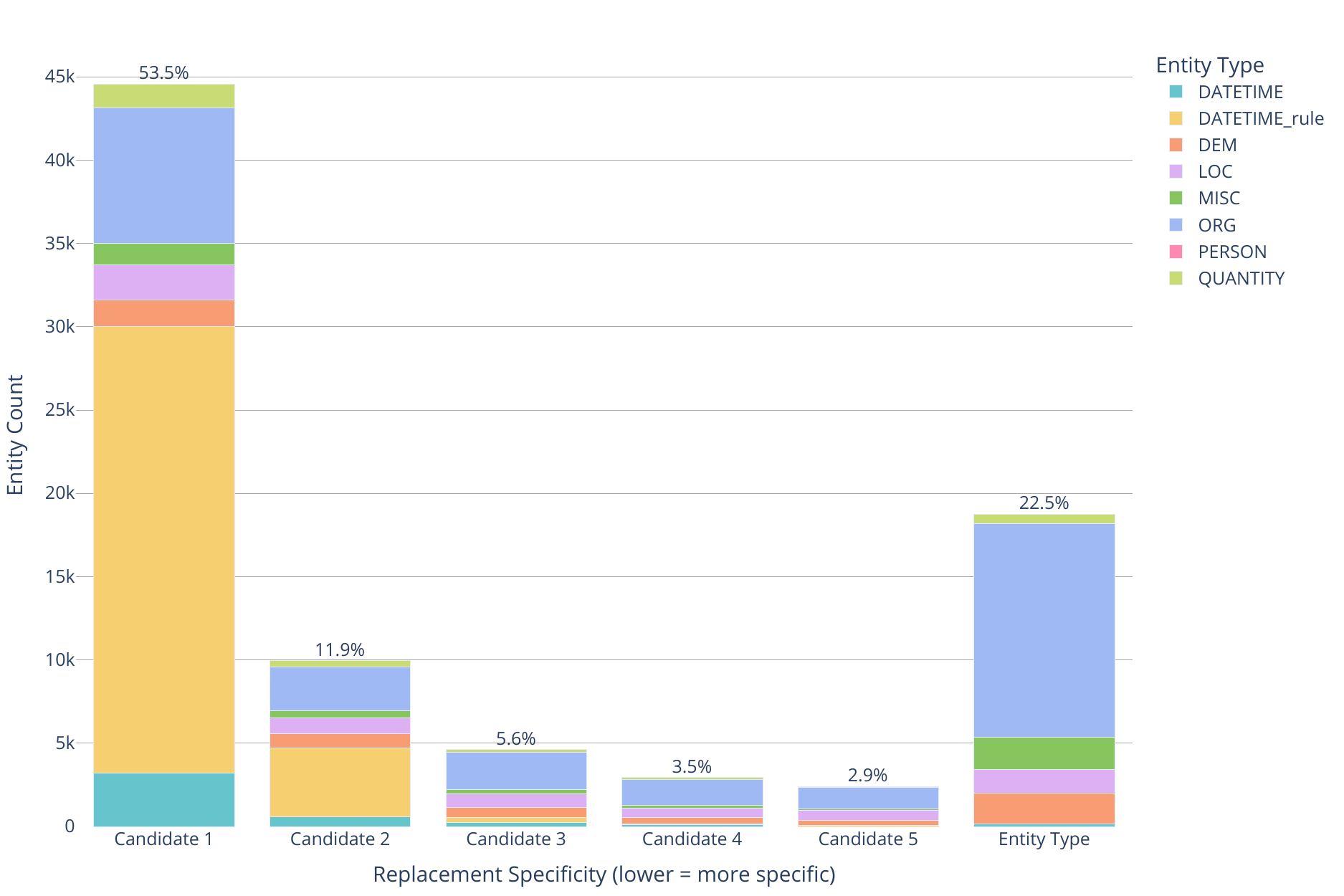}
        
        \caption{Frequency of INTACT's replacement choice for TAB. \textit{Entity Type} is selected when all five candidates are guessed by the inference attacks, leading to a default strategy of replacing the text span by its entity type label (e.g.~\ent{PERSON}).}
        \label{fig:repl_spec_DEV}
\end{figure}

Figure \ref{fig:repl_spec_DEV} shows the frequency at which each of the five replacement candidates is selected by the INTACT approach. The candidates are sorted by the LLM by order of abstraction, from the most specific to the most generic/abstract.
In 53\% of the text spans from the TAB corpus, the most specific (first) replacement candidate is chosen\footnote{For \ent{DATETIME} entities, the frequent selection of Candidate 1 and 2 can be attributed, in part, to the stricter definition of what constitutes a ``successful'' guess, where only complete lemma overlap was considered as guessed.}. In about 22\% of the text spans , all five replacement candidates were guessed by the LLM and the entity label was thus used to replace the text span. This default replacement strategy was particularly common for entities labeled as \ent{ORG} and \ent{MISC}, amounting to almost half of text spans ending up in this default case, as detailed in Table \ref{tbl:entity_type_masking_stats}.

The experiments were run locally on a single NVIDIA RTX 4000 SFF Ada GPU with 20GB of memory. The candidate generation step took approximately 3 hours for the 127 documents containing over 7,300 entity mentions in the TAB test set, while inference attacks took considerably longer (approximately 15-20 times more) with the smaller, quantized open LLMs tested. It is worth noting that LLM calls for replacement selection are performed for each replacement candidate for each entity mention based on a whole document. In contrast, generation is performed once per each unique entity.

\subsection{Utility evaluation}
\label{sec:exp_results_utility}

Utility preservation is examined from both intrinsic and extrinsic viewpoints, as detailed below.

\begin{figure}[t!]
\includegraphics[width=13.7cm]{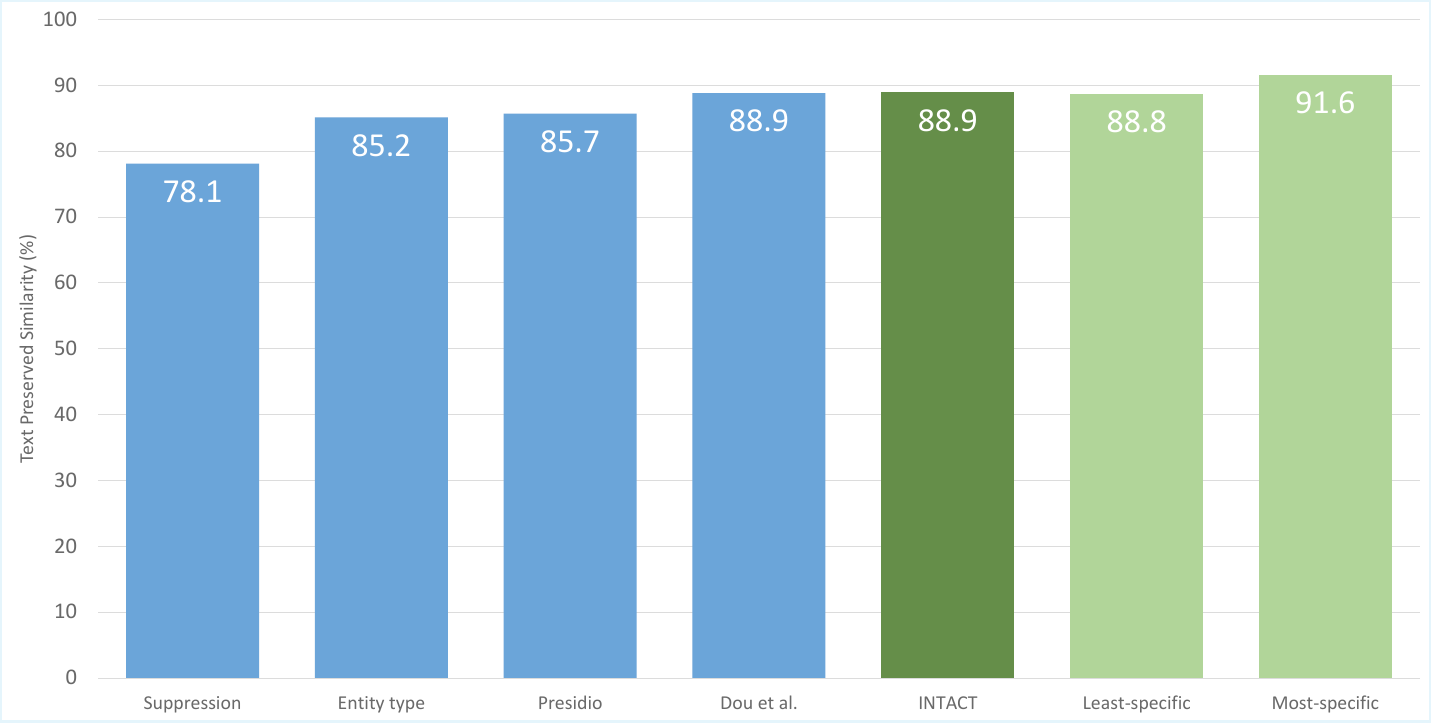}
\caption{Text Preserved Similarity (TPS) for the different replacement strategies (higher is better). The three last results, shown in green, all leverage the Mistral 7B instruct model.}
\label{figure:tps}
\end{figure}

\paragraph{Intrinsic evaluation} We first assess the preservation of the document's semantic content by comparing the original documents and their protected versions using the application-independent TPS metric proposed in Section \ref{sec:eval_utility}\footnote{To determine the value for the hyper-parameter $N$ specifying the spacing between text spans replaced by \begin{scriptsize}\texttt{[MASK]}\end{scriptsize} tokens (see Section \ref{sec:ic}), we tested multiple values (\emph{i.e.}, $N\in[2,4,6,8]$), and selected $N=6$ as a trade-off between runtime and context size.} Figure \ref{figure:tps} depicts the TPS values obtained for the different strategies. Since all of them protected the same text spans (manually annotated in TAB), discrepancies in TPS scores reflect the varying semantic similarities between the original text spans and their replacements.

The lowest utility corresponds to the suppression strategy, whereas utility figures strictly increase as we employ more specific/utility-preserving replacements, that is: entity types $\rightarrow$ Presidio replacements $\rightarrow$ least-specific replacements $\rightarrow$ Dou et al. replacements $\rightarrow$ INTACT replacements $\rightarrow$ most-specific replacements. The average semantic similarities for the different replacements were 0.33, 0.38, 0.49, 0.49, 0.51, and 0.62, respectively. 

As mentioned in Section \ref{sec:exp_dataset}, about 11\% of the text spans in TAB were marked by the annotators as expressing direct or indirect PII. The suppression of those PII-related text spans causes a 22\% drop in TPS, suggesting these spans are, on average, twice as informative as the non-protected spans. As shown in Figure \ref{fig:repl_spec_DEV}, the two most common replacement candidates selected by INTACT are respectively the first candidate of the sorted list of replacements or the default entity type, which explains why its TPS falls between that of entity replacements (85\%) and the most-specific replacements (92\%). In general, replacement-based methods show TPS differences less than 5\%. This is likely due to the fact that they share a common set of text spans as replacement candidates, with the broadest generalization (entity type) already resulting in TPS=85\%.

\paragraph{Extrinsic evaluation} To further validate the results, we measured to which extent document's utility was preserved by the different strategies on a downstream task: document clustering. To quantify the task-oriented utility preservation, we compared the clusters obtained from the original documents (viewed as the ground truth), with those resulting from the protected documents under the different strategies. Utility preservation was then computed as the similarity between both clusterings, as similar clusters suggest that equally similar conclusions can be reached from the analysis of the protected documents.
Document clustering, being a broad and unsupervised task based on the similarities of documents, serves as a good proxy for how the sanitized texts could perform in a variety of applications. It is particularly useful because it does not rely on predefined labels, thereby reflecting how well semantic and topical structures are preserved post-sanitization.

The clustering method followed the approach proposed in \cite{subakti-bert-clustering}, which relies on BERT \cite{devlin-bert}\footnote{More specifically the \textit{bert-base-cased} model, see \url{https://huggingface.co/google-bert/bert-base-cased}.} to generate document embeddings by averaging the embeddings over all individual tokens. Once the document embeddings were obtained, they were clustered using the K-means++ algorithm \cite{arthur-k-means++} with $K=4$. The value of $K$ was selected based on the optimal inertia achieved when clustering the original documents. To mitigate the effect of the random initialization, the clustering for each document set was repeated 50 times, choosing the results with the lowest inertia. To quantify the similarity between clustering results, we used the \textit{Normalized Mutual Information} (NMI) \cite{ana-2003-nmi}. For NMI calculation, the average of 5 runs of the aforementioned clustering process was computed. Each clustering was therefore conducted $50 \times 5$ times to minimize the impact of random initializations.

\begin{figure}[t!]
\includegraphics[width=13.7cm]{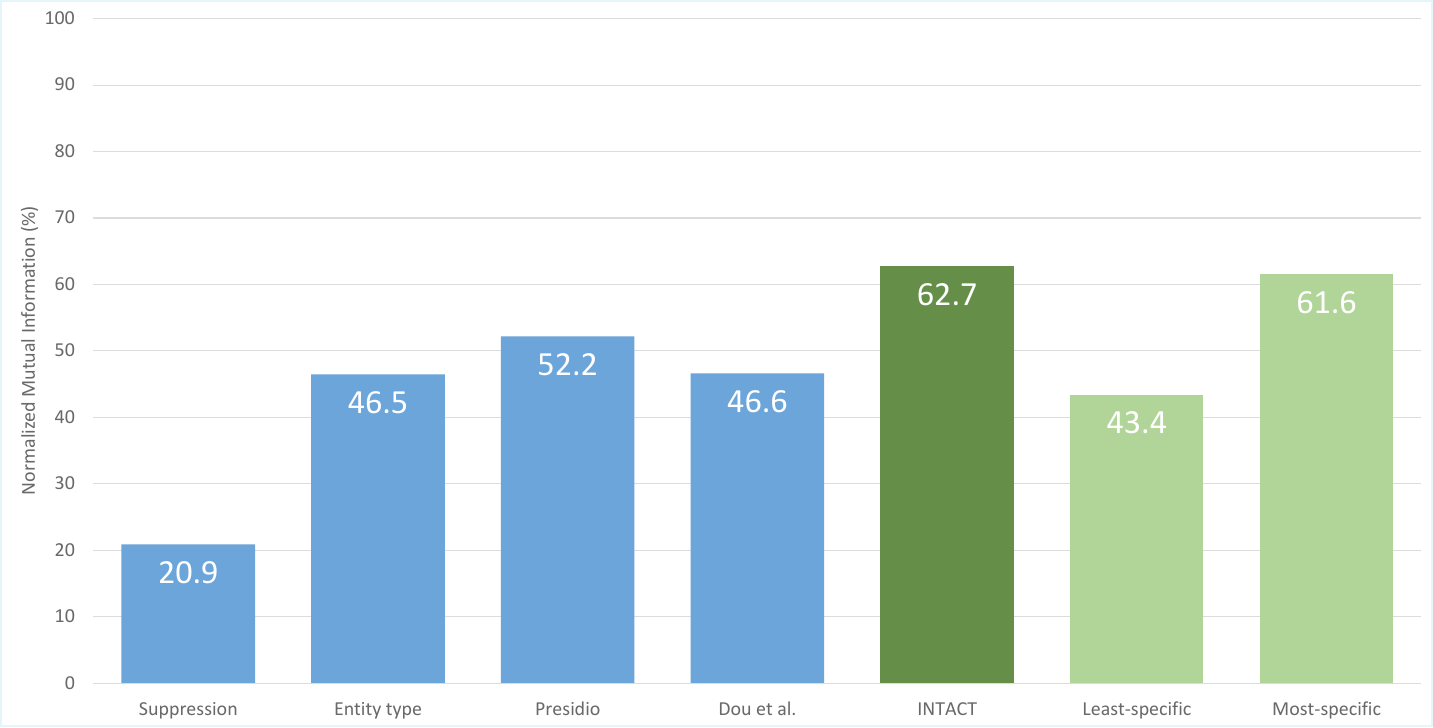}
\caption{Normalized Mutual Information (NMI) for the clustering results obtained after applying the various replacement strategies (higher is better). The three last results (in green) leverage the Mistral 7B instruct model.}
\label{figure:nmi}
\end{figure}

Clustering-based utility evaluation for the considered strategies are shown in Figure \ref{figure:nmi}. If we compare the results of this clustering-based utility evaluation with TPS, we obtain a strong Pearson correlation $r=0.864$ ($p$-$value=0.012$). This shows the effectiveness of TPS as an \textit{a priori} estimator of the analytical utility of the protected documents in downstream tasks. Minor differences between approaches stem from TPS assessing semantic similarity in isolation, without considering the broader linguistic context. A term may be semantically appropriate alone but poorly integrated in its surrounding context. While downstream results suggest minimal impact on the overall semantic content, TPS scores does not directly account for readability and stylistic nuances.

Method-level differences for NMI offer additional insights into how various approaches impact empirical data utility. For instance, INTACT achieves the highest NMI, even higher than the most-specific replacements, while their TPS remains lower. Similarly, entity replacements obtained a higher NMI than least-specific replacements, while the opposite holds for TPS. A common factor in both cases is the use of entity types. Although generated replacements are more similar to the original text spans when examined individually (\emph{i.e.}, higher TPS), some tend to distort the general document content more than entity types (\emph{i.e.}, lower NMI). Such distortion likely arises when a generalization leads to overly broad replacements that disrupt the linguistic register and structure of the document (as exemplified by the three least-specific replacements in Example \ref{fig:intact_overview}), or even unfaithfully abstract its initial content. The strategy of using entity types when all generated replacements are guessed (which is frequent, as shown in Figure \ref{fig:repl_spec_DEV}) actually resulted in a NMI that exceeded that of the most-specific replacements. It seems that, when the language model fails to generate replacements that sufficiently protect privacy, it also struggles to produce options that effectively preserve utility.

Discrepancy between TPS and NMI is more pronounced with Dou et al. and least-specific replacements, which achieved a TPS nearly identical to INTACT but resulted in a much lower NMI. The replacements obtained with these methods are often lengthy sentences that, while resembling the original span (\emph{i.e.}, high TPS), tend to interfere more with the general content of the text (and thus yielding a low NMI).

Presidio's synthetic replacements achieved a high NMI, outperforming both entity-based and least-specific replacements, likely due to their better integration within the text. Nonetheless, even when using a more advanced model than the one used to generate our replacements (\emph{i.e.}, GPT-3.5-turbo instead of Mistral-7B), its non-truthful replacements still yielded a lower NMI compared to our most-specific and INTACT's selected replacements.

\subsection{Privacy evaluation}
\label{sec:exp_results_privacy}
We evaluated the residual disclosure risk of the strategies by using the TRIR metric introduced in Section \ref{sec:eval_privacy}. 

To this end, we leveraged the ready-to-use implementation available in the corresponding GitHub repository \footnote{\url{https://github.com/BenetManzanaresSalor/TextRe-Identification}} with the default hyperparameters, including the use of \textit{distilbert-base-uncased}\footnote{\url{https://huggingface.co/distilbert/distilbert-base-uncased}} \cite{sanh-2019-distilbert} as language model. This model is fine-tuned for the re-identification task, what allows to obtain results competitive with those of pre-trained LLMs \cite{bucher-2024-finetuning-lms} without extensive computational resources. For instance, fine-tuning on an Nvidia RTX 4080 took just 36 minutes.

TRIR rests on the definition of a body of \emph{background knowledge} from which re-identification risk is assessed. To define this background knowledge, we used a corpus of 13,759 rulings published on the ECHR website. Since the annotated court decisions in TAB were drawn from this same source, the full corpus represents a strict superset of the TAB documents. However, assuming that an adversary has access to the entire corpus is unrealistic --such an adversary would already possess the original documents and thus have no incentive to re-identify their sanitized versions. It is also unrealistic to assume the adversary’s knowledge includes only the individuals in the protected dataset, as adversaries typically lack precise knowledge of the dataset’s population. Given these considerations, we defined the background knowledge as a random superset of TAB, comprising twice the number of individuals/documents (i.e., $1,268 \times 2 = 2,536$), and included only the first 10\% of non-\textit{Procedure} text from each ruling.

\begin{figure}[t!]
\includegraphics[width=13.7cm]{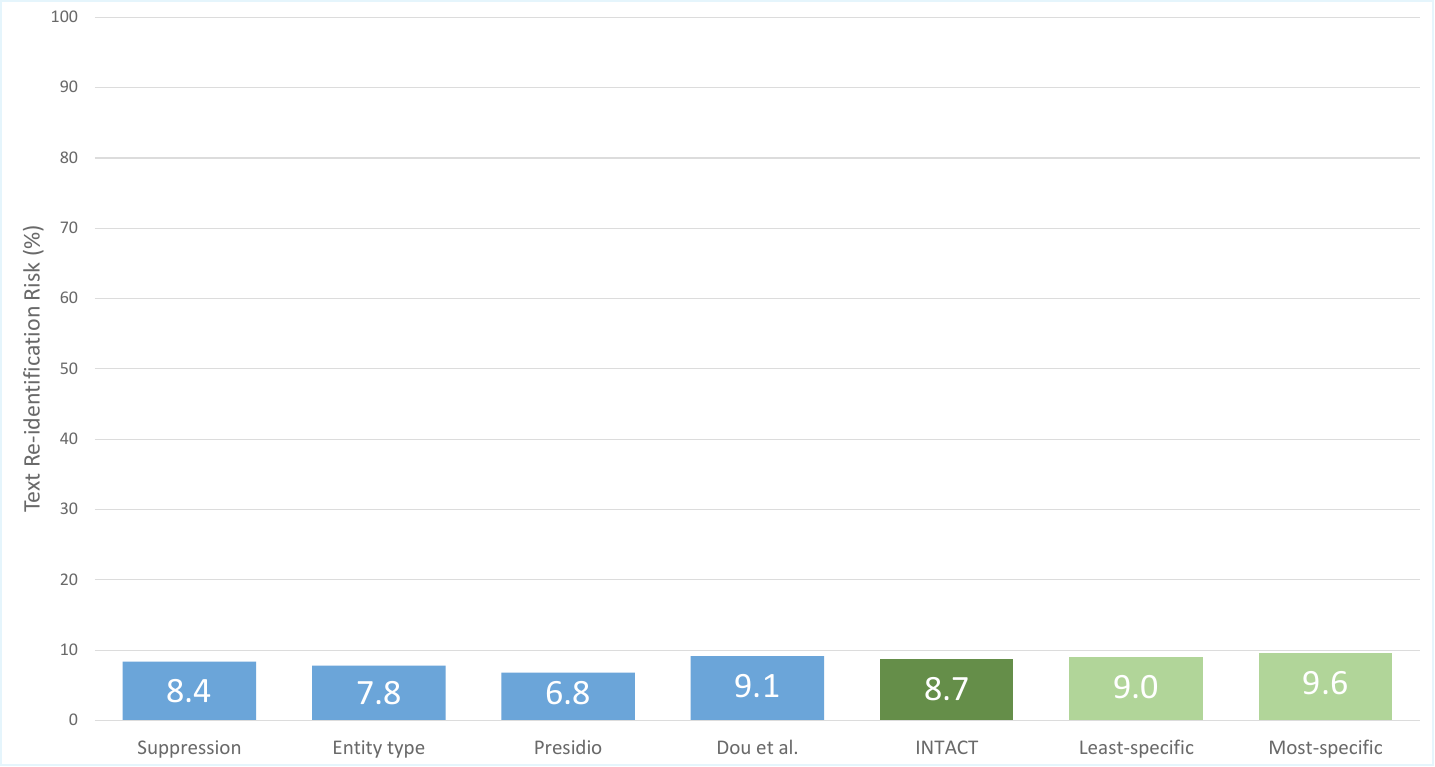}
\caption{Text Re-identification Risk (TRIR) obtained with the various replacement strategies (lower is better). Results for our approach (marked in green) leveraged the Mistral 7B instruct model.}
\label{figure:trir}
\end{figure}

The TRIR for each strategy is reported in Figure \ref{figure:trir}. Notice that, as any \emph{ex post} privacy metric, TRIR measures \emph{re-identification risk} rather than \emph{actual re-identification}. Subsequently, the $\approx10\%$ figures resulting from all methods do not mean that 10\% documents have been unequivocally re-identified, but that attackers have a 10\% chance of accurately guessing the protected individual, as they cannot verify the correctness of their predictions.

As expected, the replacements with the most-specific LLM generalizations yield the highest (\emph{i.e.}~worst) TRIR scores. The difference between this approach and the one yielding the lowest TRIR score (the suppression strategy) is, however, relatively small (2.2 percent points). In other words, the reliance on informative replacements (generalizations) makes it possible to substantially improve utility without substantially impairing privacy with respect to the strictest protection, \emph{i.e.}, suppression. INTACT performed particularly well on this balance, achieving a relatively low TRIR overall (second lowest when considering only generated replacements), while delivering the second-highest TPS and the highest NMI. Compared to both the least-specific and most-specific replacements, INTACT delivered lower TRIR alongside comparable or superior TPS and NMI.

On the other hand, Presidio's and entity replacements incurred the lowest risks because of the non-truthful replacements of the former (see Section \ref{ssec:truth}), and the abstract replacements of the latter tended to confuse the re-identification model. This reduced re-identification accuracy even more than suppression, because the re-identification model relies on concepts found in the background knowledge of the individuals targeted for re-identification. While suppression removes part of that exploitable information, Presidio's synthetic and entity replacements substitute it by \emph{new} information that is rarely found in any background knowledge.

The scenario differs for truthful replacements --like INTACT and Dou et al.--, where part of the meaning of the original concept is preserved, making it possible to connect it with the corresponding individual's background knowledge, which slightly increases TRIR. However, regardless of the actual TRIR, the use of non-truthful replacements, as is the case for e.g.~Presidio, may allow an attacker to differentiate between protected and non-protected terms, thereby providing an advantage that can be exploited to conduct case-specific attacks focused on inferring only the protected terms. Furthermore, a document bearing explicit markers of text sanitization (such as suppressions or replacements with generic placeholders) raises awareness of the document’s sensitivity and may thereby attract more attention as a target for attacks \cite{Bier2009}.

Dou et al. replacements achieved the second-highest TRIR, closely matching that of the least-specific replacements. In fact, this similarity extends consistently across all three metrics --TPS, NMI, and TRIR--, which suggests that this method is similar to selecting the least-specific candidate generated by our approach.

\subsection{Truthfulness and abstraction level}
\label{ssec:truth}

An important characteristic of the INTACT approach is its emphasis on producing \emph{truth-preserving} replacements whose semantic content subsumes the content of the original text span. Such replacements should therefore correspond to generalizations that encompass and abstract away the original text span. However, the truthfulness and abstraction level of the replacements are nuances that are challenging to evaluate automatically in a reliable manner. Their impact in combination is reflected in the results of our automatic metrics, but these do not provide explicit information about replacement truthfulness and abstraction level separately.

To address this, we conducted a manual evaluation focused on the truthfulness and abstraction level of the selected replacements. This evaluation was performed on a subset of TAB documents, and comprised 600 instances, each consisting of a PII span mention with 70 characters preceding and following it as context.
For each document, we randomly sampled a mention for each entity and then selected a random subset of all these sampled mentions, using an equal size of 100 instances per each entity type, excluding the \ent{PERSON} and \ent{CODE} categories. The instances were rated by a human evaluator with NLP background, who was also allowed to look up concepts online whenever necessary. 

The selected replacements were assessed using two criteria: abstraction level and truthfulness.   
A replacement was considered more abstract if it constituted a \textit{hypernym} or provided fewer details than the original PII span.
A replacement was truthful if it retained the same overall content of the original span. When approximate equivalence could be assumed in the context of the span (\emph{e.g.},~\textit{28 days} and \textit{1 month}), it was counted as truth-preserving. Partially true spans (\emph{e.g.}, containing two dates) were not considered truthful.

To determine the statistical reliability of this human evaluation, we measured inter-rater agreement using assessments from an additional human rater. We randomly sampled a total of 148 instances for this purpose, equally distributed across the three methods and the six sanitized entity types used. The two raters agreed on 90\% of specificity evaluations and 93\% of truthfulness evaluations. Cohen's $\kappa$ \cite{cohen1960coefficient} which accounts for chance agreement, was 0.76 for specificity and 0.84 for truthfulness, corresponding to substantial and almost perfect agreement, respectively \cite{landis1977measurement}. 
Differences in annotators’ specificity judgments were observed in cases where proper names in the original span were replaced with common nouns conveying the same meaning (\emph{e.g.}, \textit{Widow’s Bereavement Allowance} replaced by \textit{a specific allowance for widows}). Some disagreements regarding truthfulness likely arose from the subjective interpretation of replacements with vague wording (\emph{e.g.}, \textit{late evening} for \textit{7.30 p.m.}) and from numeric information where the original and the replacement spans differred only slightly and were thus considered approximately equivalent (truthful) in some cases (\emph{e.g.}, replacing \textit{10.45} with \textit{10.46}). Another potential source of disagreement was polysemy, as in the case of whether a \textit{facility for detention} is a truthful replacement for \textit{Oakington}, which refers to both a village and an immigration detention center in the United Kingdom.

Table \ref{tab:truthfulness} presents the manual evaluation results for Presidio, Dou et al.'s model \citep{dou-etal-2024-reducing} and INTACT. According to this manual evaluation, 81.3\% of the INTACT replacements were deemed as less specific and 93.2\% as truthful.
In contrast, Presidio's manual evaluation reports 36.2\% less specific and just 19.7\% truthful replacements. Dou et al.'s abstraction model also produced mostly more abstract and mostly truthful replacements (78.3\% and 95.2\% respectively). However, replacements that were truthful but not more abstract amounted to 20.2\% for Dou et al., which is 5\% higher than for INTACT, where the iterative inference attacks are meant to prevent the use of too specific replacements. This corroborates the findings from the TRIR-based automatic evaluation results (Figure \ref{figure:trir}) which show a higher privacy risk posed by Dou et al. than INTACT.
 
\begin{table}[t!]
\caption{Human evaluation of replacements on the \emph{truthfulness} and \emph{abstraction} level for three selected sanitization methods: Presidio, Dou et al and INTACT (our approach).}
\vspace{2mm}
\label{tab:truthfulness}
\centering
\begin{tabular}{lp{1mm}rrrr} \toprule
Approach: &&  \multicolumn{2}{c}{Not-truthful}  & \multicolumn{2}{c}{Truthful}\\ \midrule
&& \begin{small}Same level of\end{small} & \begin{small}More\end{small} & \begin{small}Same level of\end{small} & \begin{small}More\end{small} \\[-1mm]
&& \begin{small}abstraction\end{small} & \begin{small}abstract\end{small} & \begin{small}abstraction\end{small} & \begin{small}abstract\end{small} \\ \midrule
Presidio \citep{MsPresidio} && 61.8\%   & 18.5\% & 2\%   & 17.7\%  \\
Dou et al. \citep{dou-etal-2024-reducing} &&  1.5 \%  & 3.3\%     & 20.2\%     & 75\%  \\
INTACT  && 3.5\% & 3.3\%  & 15.2\%    & 78\%  \\  \bottomrule
\end{tabular} 
\vspace{4mm}
\end{table}

The results from Table \ref{tab:truthfulness} show that smaller, open LLMs are well suited to finding generalized replacements in terms of abstraction level and, especially, truthfulness, even without specific fine-tuning for this task. For both INTACT and Dou et al., however, the generation of more abstract replacements was somewhat more challenging for the \ent{DEM} and the \ent{MISC} categories, where text spans are often more specialized (typically medical and legal) terms or foreign, possibly culture-specific words. The LLMs showed some tendency to produce definitions, explanations or paraphrases for these terms (\emph{e.g.}, \textit{a report on an individual's health status from a medical institution} as a replacement for the Swedish medical term \textit{epikris}).

In some cases, none of the suggestions were generalizations, \emph{e.g.}, \textit{unemployed} resulted in the INTACT replacements: \textit{Out of work, Jobless, Inactive in the labor force, Economically inactive, Without employment}. 
We also noticed that the truthful and more abstract replacements obtained with Dou et al. were often more generic (\emph{e.g.}, \textit{in a specific period} for replacing a \ent{DATETIME} entity) than the ones for INTACT. This type of semantic loss is likely one of the reasons why INTACT outperformed Dou et al. in the clustering-based utility evaluation results in Figure \ref{figure:nmi}.

In comparison, Presidio's replacements were much less consistent, with a vast majority of them being non-truthful and with arbitrary degrees of abstraction. The lack of truthfulness and consistency is expected to significantly distort the document's meaning. Moreover, we noticed a substitution bias towards US-related sources, with the nationality of defendants in the TAB corpus changed from Turkish, Polish and British to Brazilian, Canadian and Mexican. Even though these distortions contribute to preventing re-identification (as shown in Section \ref{sec:exp_results_privacy}), they also seriously hamper secondary uses, as it might invalidate conclusions drawn from per-document analysis.

The results from this manual evaluation therefore show that the two-stage approach of INTACT leads to replacements that are truth-preserving in the vast majority of cases, and also typically more abstract than the original text span. As non-truthful replacements are expected to impair utility, while replacements at the same level of abstraction may compromise privacy protection, those results are in line with the ones obtained with the automated evaluation results of the previous sections.

\subsection{Generalizability to other LLMs}
\label{ssec:exp_results_generalizability}

To empirically validate the generalizability of our method to other LLMs, we applied it to the TAB test set using another open-weights model, namely Llama 3.1 8B Instruct\footnote{\url{https://huggingface.co/meta-llama/Llama-3.1-8B-Instruct}} \cite{grattafiori2024llama}. Although both Llama 3.1 and Mistral 7B are transformer-based, they have some architectural differences, with Mistral 7B emphasizing efficiency through optimized attention mechanisms such as Sliding Window Attention. However, while Mistral 7B is primarily designed for English, Llama 3.1 offers strong performance in seven languages.

For the Llama 3.1 experiments, the prompts, parameter settings, and infrastructure were identical to those used in the Mistral-based experiments described in Section \ref{sec:exp_tab}. We evaluated the outcomes using the same automatic utility and privacy metrics as in Sections \ref{sec:exp_results_utility} and \ref{sec:exp_results_privacy}, with one change: the background knowledge used by TRIR was adapted to this specific data subset. In Section \ref{sec:exp_results_privacy}, the background knowledge was a superset twice the size of the protected data. To replicate an equivalent level of adversarial robustness, we used the background knowledge from TAB’s test and dev sets. Since both sets have the same size (\emph{i.e.}, 127 individuals/documents), this forms a superset twice the size of the protected data, maintaining a comparable threat model.
Table \ref{tab:generalizability} reports the results for both Mistral and Llama LLMs, along with the suppression-based baseline and related work for reference on the TAB test set.

\begin{table}[ht!]
\caption{Utility and privacy evaluation for LLM generalizability on the TAB test subset with Llama 3.1 8B. ($\uparrow$ for higher is better, $\downarrow$ for lower is better)}
\vspace{2mm}
\label{tab:generalizability}
\centering
\begin{tabular}{lp{1mm}rrrr} \toprule
Method && TPS$\uparrow$ & NMI$\uparrow$ & TRIR$\downarrow$ \\ \midrule
Suppression && 77.7\%   & 30.2\% & 11.0\% \\
Entity type && 84.8\%   & 64.2\% & 13.4\% \\
Presidio~\citep{MsPresidio} && 85.8\%   & 66.0\% & 11.8\% \\
Dou et al.~\citep{dou-etal-2024-reducing} && 88.4\%  & 58.7\% & 15.0\% \\
Mistral INTACT  && 88.5\% & 80.0\%  & 12.6\% \\
Mistral least-specific  && 88.5\% & 57.7\%  & 13.4\%  \\
Mistral most-specific  && 91.3\% & 89.9\%  & 14.2\% \\
Llama INTACT  && 88.7\% & 86.6\% & 13.4\%  \\
Llama least-specific  && 87.4\% & 56.6\%  & 13.4\%  \\
Llama most-specific  && 90.4\% & 88.7\%  & 15.0\% \\  \bottomrule
\end{tabular} 
\vspace{4mm}
\end{table}

Results demonstrate that INTACT performs consistently across both LLMs, confirming its generalizability. Llama 3.1 achieves slightly better sanitization performance than Mistral 7B, with a NMI 6\% higher while maintaining TPS and obtaining a only 1\% higher TRIR --which, for this 127 individuals set, implies that only one additional individual has been correctly re-identified. In addition, the superior privacy and utility trade-off observed for INTACT in previous sections also holds here: INTACT always achieves the highest NMI, equal or higher TPS, and equal or lower TRIR compared to related work, exceeding suppression-based masking by only 3\%.
For least-specific and most-specific replacements, Llama’s scores are about one percentage point below Mistral’s. This suggests that Llama produced slightly inferior lists of sorted replacement candidates in the generation step, but the gap is small compared to the differences with other methods. Furthermore, it highlights the impact of the Llama-based selection step, which, despite starting from worse candidate lists, still produced outcomes on par with --or better in the case of NMI-- than Mistral-based INTACT. Mistral and Llama differ in how often the entity type was selected instead of the model-generated replacements. For Llama, the entity type was chosen \emph{ca.} 5\% less frequently, with a corresponding increase in the selection of the second most-specific candidate (selected for 12.8\% of entity mentions for Mistral vs. 17.8\% for Llama, see Figures \ref{fig:repl_spec_test_mistral} and \ref{fig:repl_spec_test_llama} in \ref{sec:apx_sanitization}).

\section{Discussion}
\label{sec:discussion}
The evaluation results from the previous sections corroborate that the identified gap in the literature -- namely, the absence of a principled method for generating truthful and utility-preserving replacements for sensitive text spans -- is effectively addressed by the proposed INTACT approach.

Our utility analysis shows that while INTACT performs on par with related work in intrinsic evaluation (TPS), it substantially outperforms it in extrinsic evaluation, achieving over a 10\% improvement in NMI on TAB. This extrinsic metric is particularly relevant for practitioners, as it reflects performance in a generic downstream task. Accordingly, we can claim that INTACT achieves \emph{effective} utility preservation.

This utility gain comes without a notable reduction in privacy protection. Privacy was assessed using the TRIR metric under a realistic threat model --one that specifies plausible background knowledge and computational resources (see Section~\ref{sec:exp_results_privacy}).
In this way, our adversarial robustness test aligns with Recital 26 of the GDPR\footnote{\url{https://gdpr-info.eu/recitals/no-26/}} \cite{GDPR}, which states that anonymous data falls outside the scope of the regulation and that ensuring anonymity requires considering the resources likely to be used for re-identification.
The empirical re-identification risk for INTACT was not substantially higher than that of redaction (\emph{i.e.}, suppression-based masking). Concretely, the risk figure was approximately 9\% (equivalent to $k=11$ or greater in probabilistic $k$-anonymity terms \cite{samarati-k-anonymity, probabilistic-k-anonymity}), what according to \cite{elemam-2013-guide-deid-medical} corresponds to a strong degree of anonymity, as attested by two independent health regulators \cite{european-2014-medical-regulations,canada-2019-medical-regulations}.

Our manual evaluation of truthfulness and abstraction further supports these findings: INTACT’s replacements preserved the core meaning of original spans in 93\% of cases, while avoiding simple rewrites in 85\% of cases. This indicates that the method produces genuine generalizations, which can improve utility without significantly undermining privacy.

Importantly, these results were achieved using relatively small, instruction-tuned LLMs with open weights—specifically, Mistral (7B parameters) and Llama 3.1 (8B parameters). This demonstrates that compact models can be highly effective for efficient text sanitization. The strong performance of Llama 3.1 further confirms that the method generalizes well across different model architectures, with no notable loss in either utility preservation or privacy protection. This consistency suggests that INTACT’s effectiveness is rooted in its core design principles rather than in model-specific characteristics, highlighting its robustness and broad applicability.

Overall, INTACT presents a compelling solution for data holders aiming to release de-identified documents in compliance with privacy regulations, while maintaining the semantic and analytical integrity of the data. However, integrating the proposed method into end-to-end NLP systems must take the three following aspects in consideration:
\begin{itemize}
\item To support languages beyond English, the selected LLM\footnote{This also holds for the generalization heuristics employed for \begin{scriptsize}\textsf{DATETIME}\end{scriptsize} entities.} must support the target language to accurately identify and replace sensitive spans.
\item Due to the computational costs of running inference attacks on candidate replacements, the current approach is best suited for small- to medium-size document collections. Future improvements in data handling and candidate selection strategies could significantly enhance runtime performance and reduce computational overhead.
\item Finally, like all sanitization techniques, INTACT does not offer \textit{ex ante} privacy guarantees. Consequently, an \textit{ex post} empirical privacy evaluation -- such as TRIR -- remains essential. Complementary assessments of utility preservation (\emph{e.g.}, TPS or NMI), truthfulness, and abstraction are also important to ensure the quality of the released documents. When these evaluations themselves rely on language models, it is important to ensure those support the target language; for instance, multilingual sentence-transformers with aligned cross-lingual representations can be used for TPS in multilingual contexts.
\end{itemize}

\section{Conclusions and future work}
\label{sec:conclusions}

We presented INTACT, a two-stage approach to automatically sanitize disclosive text spans in documents by leveraging instruction-tuned large language models. In contrast to previous approaches, the presented method is not limited by the narrow scope of structured knowledge bases and focuses on generating \emph{truthful} replacements, which are of outmost importance in secondary uses of the data, such as medical research, insurance or law. A key benefit of the presented method is its ability to identify the most privacy-preserving candidates by testing whether a language model can infer the original span through an inference attack. Given the advanced capabilities of LLMs as adversaries (see e.g.~\citep{Charpentier2025-wa}), this helps ensure that replacements duly conceal the personally identifiable information present in the original text spans. We ensure the replicability of our method by using open-weights LLMs, which can also be run locally to guarantee private and secure handling of sensitive data. Empirical results show that our method provides a better privacy/utility trade-off and avoids distorting the original semantic content of the document as other LLM-based approaches do.

We have also proposed Text Preserved Similarity (TPS), an automatic metric for text anonymization that captures the notion of semantics better than standard precision metrics, does not require (costly) manual annotations as ground truth and is compatible with sanitization.

In future work, we aim to enhance both the ordering of replacement candidates and their grammatical integration within sentences. Even though a grammatically correct output is specifically requested in the replacement generation prompt, some replacements occasionally produce ungrammatical phrases (\emph{e.g.}, double prepositions like \textit{in in}) if inserted, partly due to inconsistencies in the inclusion of these within the manually annotated spans of the dataset used. The fluency of the output is an additional aspect that could be assessed manually by future work. Moreover, while the replacement suggestions generally range from the most specific to the most abstract, the ordering is not always consistent. We also plan to explore additional methods for matching the original spans against those guessed by the LLM. These enhancements could potentially be implemented through additional LLM steps to further improve the quality of the sanitized text. Future studies could also investigate whether these challenges persist across other LLMs. Given that some of our findings may be limited to the specific LLMs tested, a broader comparative analysis based on a variety of model types, sizes and configurations would allow for validating the robustness and generalizability of the proposed approach and results.

Future work will also seek to address the limitation of TPS with assessing terms' similarity in isolation, mentioned in Section \ref{sec:exp_results_utility}. Tackling this limitation requires a shift from term-level to context-aware similarity metrics, such as those based on sentence-level or passage-level embeddings. While such embeddings offer a more holistic representation by capturing the semantics of entire sentences, their integration into the TPS framework poses methodological challenges. Specifically, the similarity computation would no longer be a function solely of the original and replacement spans, but would instead need to consider the full textual environment, including potentially subtle changes in meaning or tone. This shift thus requires a revised scoring that can meaningfully interpret and weigh contextual information.

A final limitation of our work is the use of a benchmark limited to English texts and focusing on the legal domain, which may constrain generalizability across domains and languages. We will therefore seek to extend the evaluation to other domains and explore the use of multilingual LLMs to assess the method’s broader applicability.

\section*{Acknowledgments}
We acknowledge support from the Norwegian Research Council (CLEANUP project (\url{http://cleanup.nr.no/}), grant nr. 308904), the Government of Catalonia (ICREA Acad\`emia Prize to D. S\'anchez, and grant 2021SGR-00115), MCIN/AEI/ 10.13039/501100011033 and ``ERDF A way of making Europe'' under grant PID2021-123637NB-I00 ``CURLING'', and the EU's NextGenerationEU/PRTR via INCIBE (project ``HERMES'' and INCIBE-URV cybersecurity chair). B. Manzanares-Salor is also supported by the Spanish Government under an FPU grant (ref. FPU23/01785).

\section*{Declaration of generative AI and AI-assisted technologies in the writing process.}

During the preparation of this work the author(s) used ChatGPT and Microsoft Co-pilot (with GPT-4o) in order to improve readability and fluency in parts of the text. After using this tool/service, the author(s) reviewed and edited the content as needed and take(s) full responsibility for the content of the published article.

\bibliographystyle{elsarticle-harv}
\bibliography{cas-refs}

\appendix

\section{LLM-based text sanitization}
\label{sec:apx_sanitization}


Table \ref{tbl:llm_examples} shows the synthetic examples used for obtaining the LLM-based replacements per semantic category.

\begin{table}
\caption{\label{tbl:llm_examples} Example sentences and replacements used in the one-shot in-context learning setup for replacement generation.}
\begin{tabular}{lp{40mm}p{62mm}}
\textbf{Category} & \textbf{Sentence context} & \textbf{Sorted replacements}\\
\toprule
\textbf{\ent{ORG}} & John Smith often volunteered in [[Sunrise Psychiatric Hospital]]. & a mental health facility, \newline a medical facility,  \newline a health-related establishment,  \newline a center for wellbeing,  \newline a public institution\\
\midrule
\textbf{\ent{DATETIME}} & Mary Smith was born on [[March 12, 1999]]. & March 1999, \newline spring 1999, \newline the first half of 1999, \newline the late 1990s, \newline the late XX century\\
\midrule
\textbf{\ent{LOC}} & John Smith often performs in [[London]]. & a large city in the UK, \newline a European capital, \newline a large island nation, \newline in the UK, \newline in Europe\\
\midrule
\textbf{\ent{QUANTITY}} & The man had [[three]] children. & between two to five, \newline a handful of, \newline a small number of, \newline over two, \newline some\\
\midrule
\textbf{\ent{DEM}} & Maria Janion was an excellent [[Polish]] scholar. & West Slavic, \newline Slavic, \newline Eastern European, \newline European, \newline Eurasian\\
\midrule
\textbf{\ent{MISC}} & John Smith served in [[World War I]]. & a military conflict in the first half of the 1900s, \newline a military conflict in the 20th century, \newline a war in Modern Times, \newline an international war, \newline an armed conflict\\
\bottomrule
\end{tabular}
\end{table}

Table \ref{tbl:entity_type_masking_stats} contains statistics over entity type replacement with the Mistral model.

\begin{table}
\centering
\caption{Statistics of using entity type as replacement in TAB in cases where the inference attack successfully guesses all LLM-generated (or rule-based) replacements  proposed.}
\begin{tabular}{crrr}
\toprule
\textbf{Entity type} & \textbf{Counts} & \textbf{Total counts} & \textbf{Percentage} \\
\midrule
\ent{DATETIME} & 175 & 4,467 & 3.92 \\
\ent{QUANTITY} & 590 & 2,709 & 21.78 \\
\ent{LOC} & 1,402 & 6,417 & 21.85 \\
\ent{DEM} & 1,842 & 5,558 & 33.14 \\
\ent{ORG} & 12,796 & 28,600 & 44.74 \\
\ent{MISC} & 1,955 & 4,267 & 45.82 \\
\bottomrule
\end{tabular}

\label{tbl:entity_type_masking_stats}
\end{table}

\FloatBarrier

\begin{figure}
        \centering
        
        \includegraphics[width=12cm]{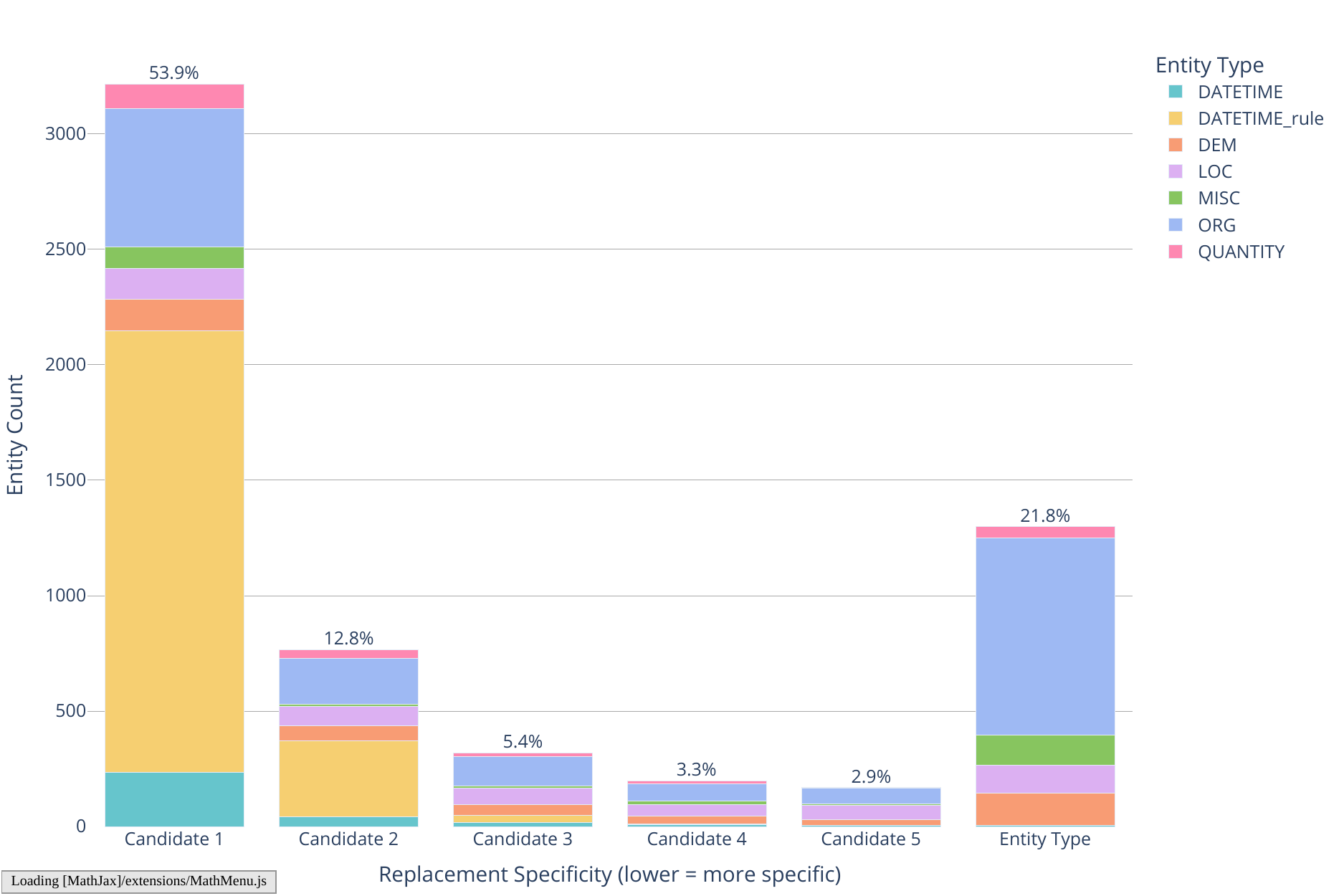}
        
        \caption{Frequency of INTACT's replacement choice for the test set of TAB with the Mistral 7B Instruct model.}
        \label{fig:repl_spec_test_mistral}
\end{figure}

\begin{figure}
        \centering
        
        \includegraphics[width=12cm]{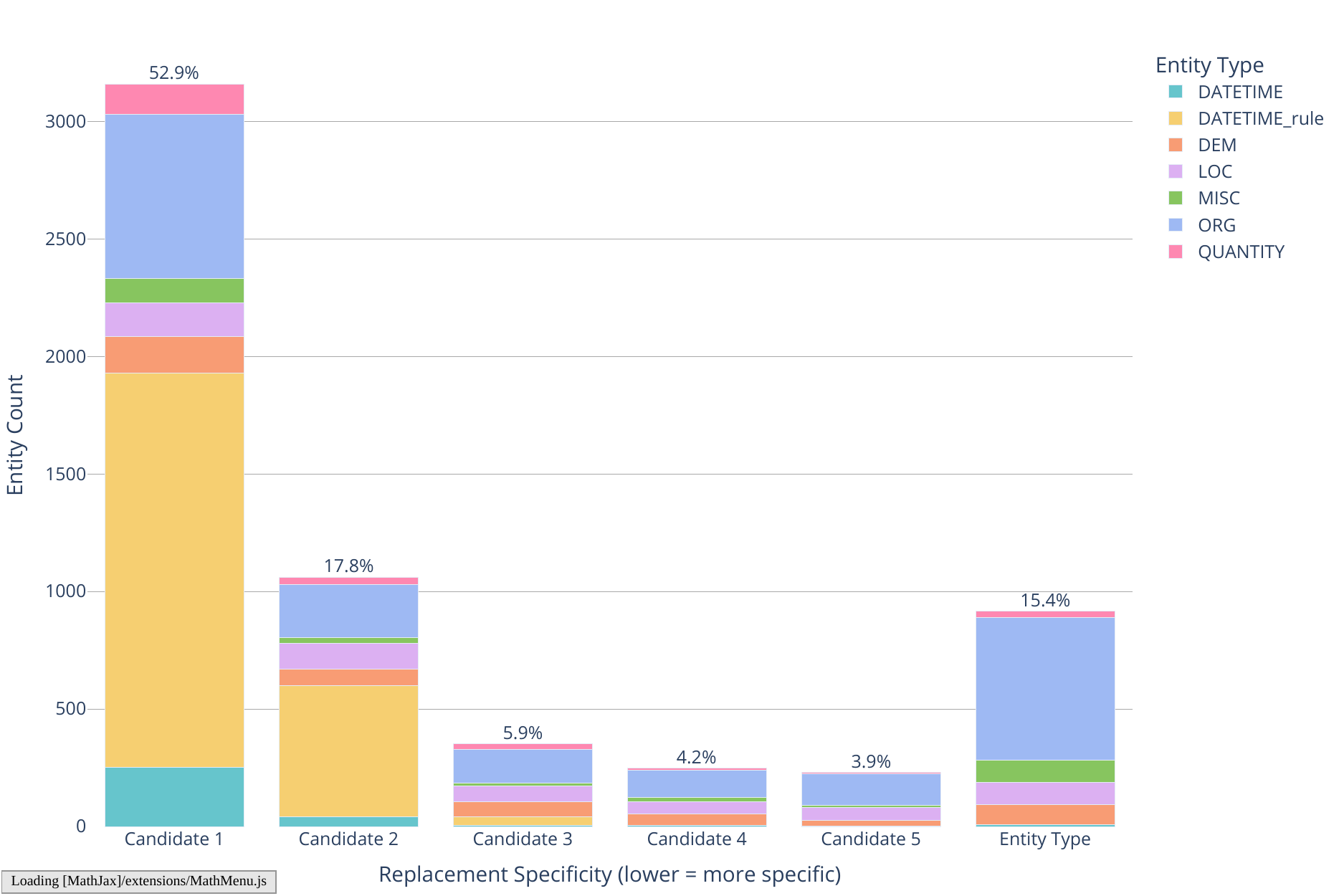}
        
        \caption{Frequency of INTACT's replacement choice for the test set of TAB with the Llama 3.1 8B Instruct model.}
        \label{fig:repl_spec_test_llama}
\end{figure}

Figures \ref{fig:repl_spec_test_mistral} and \ref{fig:repl_spec_test_llama} show how often replacement candidates with different levels of specificity were chosen by the Mistral and the Llama models respectively for the TAB test set. When all five candidates were predicted by the inference attacks, \textit{Entity Type} is chosen (\emph{e.g.},~\ent{PERSON}).

\end{document}